\documentclass[journal]{IEEEtran}

\usepackage{color,comment}
\usepackage{graphicx}
\usepackage{subfigure}
\usepackage[inline]{enumitem}
\usepackage{mathtools,amsfonts}
\usepackage[ruled]{algorithm2e}
\usepackage{lipsum}

\def\bfO{\mathbf{0}}

\def\bfL{\mathbf{L}}
\def\bfm{\mathbf{m}}
\def\bfM{\mathbf{M}}
\def\bfP{\mathbf{P}}
\def\bfs{\mathbf{s}}
\def\bft{\mathbf{t}}
\def\bfx{\mathbf{x}}

\def\bfy{\mathbf{y}}
\def\bfz{\mathbf{z}}
\def\bfmu{\boldsymbol{\mu}}

\def\c{\mathrm{c}}
\def\cm{\mathrm{cm}}
\def\ib{\mathrm{b}}
\def\d{\mathrm{d}}
\def\l{\ell}
\def\m{\mathrm{m}}
\def\mm{\mathrm{mm}}
\def\r{\mathrm{r}}
\def\s{\mathrm{s}}

\def\inFOV{\mathrm{in}}

\def\bbD{\mathbb{D}}
\def\bbN{\mathbb{N}}

\def\bbP{\mathbb{P}}
\def\bbR{\mathbb{R}}
\def\bbS{\mathbb{S}}
\def\bbT{\mathbb{T}}
\def\bbX{\mathbb{X}}

\def\calN{\mathcal{N}}
\def\calX{\mathcal{X}}

\graphicspath{{Images/}}

\SetAlFnt{\small}
\DontPrintSemicolon

\def\eqns#1{\begin{equation*}#1\end{equation*}}
\def\eqnl#1#2{\begin{equation}\label{#1}#2\end{equation}}
\def\eqnsml#1{\begin{multline*}#1\end{multline*}}

\def\eqnsa#1{\begin{equation*}\begin{aligned}#1\end{aligned}\end{equation*}}

\title{A unified approach for multi-object triangulation, tracking and camera calibration}

\author{Jeremie Houssineau, Daniel Clark, Spela Ivekovic, Chee Sing Lee and Jose Franco}

\begin{document}

\maketitle

\begin{abstract}
Object triangulation, 3-D object tracking, feature correspondence, and camera calibration are key problems for estimation from camera networks. This paper addresses these problems within a unified Bayesian framework for joint multi-object tracking and sensor registration. Given that using standard filtering approaches for state estimation from cameras is problematic, an alternative parametrisation is exploited, called disparity space. The disparity space-based approach for triangulation and object tracking is shown to be more effective than non-linear versions of the Kalman filter and particle filtering for non-rectified cameras. The approach for feature correspondence is based on the Probability Hypothesis Density (PHD) filter, and hence inherits the ability to update without explicit measurement association, to initiate new targets, and to discriminate between target and clutter. The PHD filtering approach then forms the basis of a camera calibration method from static or moving objects. Results are shown on simulated data.
\end{abstract}

\section*{Introduction}

\noindent Detection, localisation and tracking of an object's state from active sensors, such as, e.g., radar, range-finding laser and sonar, are usually determined from the sensor measurements using a stochastic filter, such as the Kalman filter \cite{kalman}, to provide statistically optimal estimates. When the use of active sensors is not possible, since it can give away the position of the sensor, passive sensors, such as cameras, are the alternative.

Calculating the distance of objects from cameras requires triangulation. The traditional means of triangulation from a pair of image observations are well known if the observations of the object are perfect, in which case the triangulated position can be calculated using the knowledge of sensor geometry \cite{Hartley2003}. However, no sensor provides perfect measurements and since the traditional methods were not designed for noisy data, these methods do not provide any notion of the quality of the triangulated estimate and can provide biased estimates when the errors in measurements are not properly considered \cite{SibleyEtAl2007}.

The objective of this paper is to describe a statistical framework for joint 3-D object state estimation and camera calibration, which considers both the geometry and the observation characteristics of the cameras. The framework presented in this paper makes use of a proxy state space, called disparity space, which allows for parts of the estimation process to be expressed in linear Gaussian form, thereby enabling the use of the Kalman filter.

The proposed framework encompasses a logical hierarchy of algorithms for estimation from noisy image measurements and addresses the following research problems: single-object triangulation, single-object tracking, multi-object triangulation, multi-object tracking, and camera calibration. Each of these research problems is motivated and discussed in Section \ref{sec:bg_related_work}.

The statistical framework is presented in a series of steps, as follows. First, the problem of triangulation from cameras is described in Section \ref{sec:triangulation_from_cameras}, followed by the description of disparity space in Section \ref{sec:disparity_space} and a discussion on the representation of object-state and object-measurement uncertainty, in the presence of disparity space, in Section \ref{sec:modelling_uncertainty}. The simplest and most constrained case of a single-object state estimation from calibrated cameras is then considered in Section \ref{sec:single_object_estimation}, followed by the case of multi-object state estimation from calibrated cameras in Section \ref{sec:multi_object_estimation}, and finally joint multi-object state estimation and camera calibration in Section \ref{sec:calibration}. Experimental results on simulated data are shown in Section \ref{sec:simulated_data}.

\section{Background and Related Work}
\label{sec:bg_related_work}

Since we are addressing a number of different problems that have previously been considered independently, we discuss each of these concepts in turn and previous approaches that have been taken in the literature. In particular, we describe the concepts of triangulation from camera measurements, the parametrisation called disparity space, tracking objects in 3-D, feature correspondence and data association and their relation to multi-object estimation, and calibration of cameras from image measurements.

\subsection{Triangulation}
\noindent Triangulation is of importance in various engineering applications, for example, surveying, navigation, metrology, astrometry, binocular vision, and target tracking. 

The vast majority of triangulation algorithms perform the estimation in 3-D directly, since we are generally interested in the object's state expressed with respect to the world co-ordinate system. However, in 3-D, due to the nonlinear nature of the perspective projection, the variance in the object position, estimated from noisy image measurements, is range-dependent~\cite{SibleyEtAl2007}, and the possible distance of the object from the cameras is unbounded, all of which makes the estimation problem very challenging.

The fact that image measurements are inherently noisy, and hence estimation from them requires statistical methodology, has been recognised by many researchers. Statistical methods for estimating the uncertainty in 3-D, such as finding the Cramer-Rao Lower Bound (CRLB), have previously been investigated~\cite{BroidaChellappa1991,ChowdhuryChellappa2005,YoungChellappa1992}, though researchers often transform the measurement or linearise the system before estimating the uncertainty, thereby losing the underlying statistical sensor characteristics in the process. There is a consensus in that the uncertainty in triangulated estimates from stereo cameras, or monocular sequences, is non-Gaussian and not trivially estimated~\cite{SibleyEtAl2007}. Furthermore, the importance of fusing the estimates given by several complementary cameras is recognised as an instrumental way of reducing the uncertainty in triangulated estimates \cite{Adiv1989,Daniilidis1993}. 

Despite these investigations, the fundamental problem of finding reliable solutions for estimation in 3-D from camera measurements remains unsolved. Our belief is that this is due to the unobservability of the object in 3-D state-space from the camera measurements. To address the shortcomings of direct 3-D estimation, we propose the use of a proxy state-space for estimation, known as {\it disparity space}.

\subsection{Disparity Space}
\noindent The concept of binocular disparity, defined by the difference in the location of an object in two images, arose from research into mammalian visual systems to reflect the horizontal separation of the left and right eyes \cite{Julesz1971}. Perception of depth is obtained in stereopsis as a consequence of this binocular disparity. The same concept is applied to problems in computer vision for extracting depth information from stereo cameras and researchers have designed algorithms for 3-D estimation from cameras by considering the disparity space as a state space~\cite{AgrawalKonolige2006,DemirdjianDarrell2002,DerpanisChang2006,Ivekovic2009,IvekovicTrucco2007}.

Most attention of the research in disparity space has focused on the case of estimation from rectified cameras (see Figure \ref{fig:disparity_space}), because, in such a scenario, estimation in disparity space has three key advantages over estimation in 3-D Euclidean space: (i) the projections into the observation space (the two image planes) are linear, (ii) the noise in the state estimate is range-independent, and (iii) the range of the estimated variable is bounded by the image size. Consequently, in disparity space, the position (but not the dynamics, as discussed in Section \ref{sec:modelling_uncertainty}) of an object can be estimated with the linear-Gaussian assumptions required for the Kalman filter update, and hence optimally and in closed form. This allows for a straightforward error analysis, for example, the computation of the Fisher information and CRLB~\cite{CoverThomas1991}. Once the CRLB in disparity space is known, the corresponding CRLB in 3-D can be derived by means of reparametrization~\cite{Clark2010}, thereby characterising the minimum variance of any 3-D state estimator from a rectified stereo pair of cameras. These results establish a practical basis for 3-D estimation theory from rectified stereo cameras that can be extended for multiple cameras with arbitrary orientations, considered in this article.

\subsection{3-D Object tracking}
\noindent 3-D object tracking refers to the problem of estimating the position {\it and} dynamics of the object. In stereo tracking, the state of a moving object in Euclidean $3$-D space is estimated from two-dimensional measurements, generated by a pair of cameras.

Object tracking requires estimation of the state of a dynamic system at each point in time, based on a sequence of noisy measurements. This definition coincides with the mathematical theory of stochastic or Bayes filtering and these terms have become synonymous in the Sensor Fusion community due to the widespread deployment of filtering techniques in practical applications. 

A stochastic filter comprises of a prediction step, based on a Markov transition that describes the motion of the object, and an update step, which updates the estimate with the Bayes rule, according to the observation characteristics of the sensor. 

An example of a stochastic filter is the Kalman filter~\cite{kalman}, which provides an optimal closed-form solution to the filtering problem when the dynamic and observation processes are linear and their corresponding noise processes are Gaussian. If the dynamic model or the observation model is non-linear and/or the motion or measurement error is non-Gaussian, then the conditions required for the Kalman filter are no longer valid. Methods for dealing with mild nonlinearities include the extended Kalman filter (EKF)~\cite{Jazwinski} or the unscented Kalman filter (UKF)~\cite{UKF}, though these still assume Gaussian motion- and measurement-noise errors.

Like in the case of triangulation, the vast majority of stereo tracking algorithms track in 3-D Euclidean space, since the operator is ultimately interested in knowing the state of the object, such as position and velocity, in the world co-ordinate system. There is a long history of using Kalman filters and their extensions to non-linear systems, such as EKF, for solving 3-D motion estimation from images (\cite{MaEtAl2006}, p437). Unfortunately, in the presence of the nonlinear observation model and range-dependent uncertainty in 3-D estimates, the usual assumption of Gaussianity in the Kalman filter leads to a poor characterisation of the posterior distribution in 3-D, particularly in the depth estimate, and hence the use of the Kalman filter and its non-linear variants will almost inevitably lead to filter divergence and poor tracking performance. This is particularly acute for targets in long-range stereo applications~\cite{SibleyEtAl2007}. 

Furthermore, the integration of these approaches into a multi-sensor system with other sensors, such as radar,  undermines the performance of the system as a whole since the probabilistic description of the object states becomes biased. Non-linear filtering approaches, specifically designed for this problem, are required in order to solve the stochastic 3-D object tracking from noisy image observations, according to the actual sensor and observation characteristics, and to successfully integrate the solution into a multi-sensor estimation framework.

\subsection{Multi-camera multi-object estimation}
\noindent Multi-camera multi-object estimation, in computer vision also referred to as the {\it feature correspondence} \cite{Hartley2003} and {\it feature tracking} problem, is a fundamental problem in estimation from images, the solution of which has a wide range of applications from object recognition, camera calibration and 3-D reconstruction to mosaicing, motion segmentation, and image morphing. It is related to the {\it data association} \cite{BS} problem in the Sensor Fusion literature: Both relate to the problem of finding the measurements which correspond to the same object that have come from different sensors, or in dynamical systems, from the same sensor at different time-steps. 

In a multi-object environment, this is a challenging task, since we may not know how many objects are in the scene, there may be many false alarms from the sensor, and there may not always be a measurement at each time-step or in each sensor. 

Methods for reducing the complexity of the problem in the sensor community usually rely on {\it gating} \cite{BS} around the object or measurement to identify possible matches, or using the {\it epipolar constraint} \cite{Hartley2003}, in the Computer Vision literature. 

Methods for data association often suffer from problems in robustness and complexity, since they usually rely on combinatorial measurement-to-object assignment approaches and do not provide a satisfactory notion of uncertainty in the multi-object state estimate. The key novelty in our approach is that we pose this problem as a multi-sensor multi-object static estimation problem, where each camera provides multiple measurements in each image frame, and the objective is to estimate the true image locations optimally. 

Recent developments in the Sensor Fusion community have enabled practitioners to overcome the computational limitations of combinatorial data association approaches by modelling the system as an integrated multi-object Bayesian estimation problem. A Bayesian solution to the multi-object filtering and estimation problem can be found with Finite Set Statistics (FISST)~\cite{Mahler2007}, a set of mathematical tools developed from point process theory, random finite sets, and stochastic geometry. 

There are a number of advantages in developing an integrated mathematical framework for multi-object detection and tracking:
(i) the number of objects and their locations can both be optimally estimated from multiple sensors;
(ii) false alarms/outliers do not need to be explicitly discarded since they will not be confirmed by the model;
(iii) the sensors are not required to provide measurements of the objects in each image and the sensor characteristics and frame rates are not required to be the same; and
(iv) advance matching of the measurements from each object is not necessary. 

The FISST approach to multi-sensor multi-object tracking has attracted significant international attention in the Sensor Fusion community due to the success of practical implementations of first-moment multi-object approximation filters, known as Probability Hypothesis Density (PHD) filters~\cite{Mahler2007,kushaclarkvo,clarkconvergence}. 

The advantage of viewing this problem as a multi-object statistical estimation problem and using the PHD filter means that, in addition to providing a rigorous mathematical foundation for multi-object estimation,
(i) there is no data association for assigning measurements to targets,
(ii) there is no need to match pairs of measurements corresponding to the same object ahead of estimation,
(iii) the objects do not need to generate observations at each time-step,
(iv) the method is robust in scenarios with false alarms,
(v) the PHD filter has a linear complexity in the number of targets and the number of measurements.
Furthermore, given a video sequence of a static scene, we can recursively apply the multi-object Bayes update on image measurements, using disparity space, and reparametrise the state estimates into 3-D, which makes the proposed approach directly extendible to the stochastic triangulation of multiple objects in cluttered environments.

\subsection{Camera Calibration}
\noindent Camera calibration refers to the estimation of the parameters of the imaging process, such that when two or more views of the same scene are available, the original $3$-D scene and its dimensions can be reconstructed by solving an inverse problem. How accurately the original scene can be reconstructed depends on the number of parameters that can be estimated and consequently different calibration methods exist. If some ground-truth knowledge about the scene is provided, e.g., a calibration object with known Euclidean $3$-D coordinates, the Euclidean calibration can be performed directly \cite{Tsai1987}. Alternatively, the so called {\it stratified approach} is used \cite{SvobodaEtAl2005}, which gradually refines the calibration from projective to Euclidean. 

In practice, a calibration object is not always available and hence the stratified approach, which relies only on the information extracted from the images, is more appropriate. Projective calibration is usually achieved by structure-from-motion techniques \cite{TomasiKanade1992} which unrealistically assume perfect knowledge of measurement correspondences as an input to the calibration process. This in turn means that such projective calibration implicitly assumes that the estimated correspondences were updated with the correct measurements and the corresponding points are known in at least a certain number of images. The possibility of incorrect data association or correspondence is not considered as such cases are pruned from the input data and similarly, the possibility of incorrect estimation of the number of correspondences is also not considered. As a consequence, useful information is removed from the input data before the calibration process even begins.

To remove the dependency of the calibration method on perfect input data, the calibration can instead be formulated as an extension of the multi-object stochastic estimation problem, discussed in the previous section. In fact, given that the projective camera calibration relies on information obtained from the multi-object state estimation, estimating the multi-object state of an {\it uncalibrated} dynamic system is inherently suboptimal if the camera parameters are not estimated as a part of the same process. 

We propose to address this problem as a doubly-stochastic inference problem~\cite{Swain2010first}, where the measurements are conditioned on the multiple-object locations, that are in turn conditioned on the relative camera orientations. A similar method using Random Finite Sets has been developed for the related problem of Simultaneous Localisation and Mapping (SLAM) for autonomous robot navigation~\cite{VoSLAM,Lee2013,Lee2014}, where each object measurement contributes both to a feature in the world and self-localisation of the vehicle.

\section{Triangulation from cameras}
\label{sec:triangulation_from_cameras}

\begin{figure}
\footnotesize
\def\svgwidth{0.97\columnwidth}
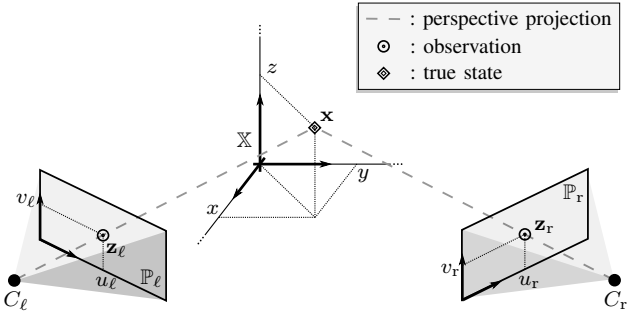
\caption{Two cameras $C_{\l}$ and $C_{\r}$ observing the same point $\bfx \in \bbX$.}
\label{fig:triangulation}
\end{figure}

The fundamental estimation problem underlying all of the algorithms presented in this paper is triangulation from a pair of cameras, illustrated in Figure \ref{fig:triangulation}. A point $\bfx$ in the real world $\bbX = \bbR^3$ is projected onto the left and right camera image planes $\bbP_{\l}$ and $\bbP_{\r}$ and its respective projections are denoted $\bfz_{\l}$ and $\bfz_{\r}$. Triangulation can then be formulated as the process of recovering the point $\bfx$ from its projections $\bfz_{\l}$ and $\bfz_{\r}$. For this purpose, the relation between the real world $\bbX$ and the image planes $\bbP_{\l}$ and $\bbP_{\r}$ must be formulated. Such a formulation can be made easier by using the concepts of projective geometry \cite{Coxeter2003}, as described next.

A point $\bfx = (x,y)^T$ in $\bbR^2$ is represented by any triple $\bar{\bfx} = (\alpha x,\alpha y,\alpha)^T$ with $\alpha \in \bbR \setminus \{0\}$, and any such triple is referred to as the \emph{homogeneous coordinates} of the point $\bfx$. A general perspective projection is a linear transformation in homogeneous coordinates, represented by an $(n-1) \times n$ matrix, where $n$ is the dimension of the original projective space. Henceforth, projective equivalents of spaces and points will be denoted with a bar. A perspective projection matrix relates the homogeneous point $\bar{\bfx} = (x,y,z,1)^T$ in $\bar{\bbX}$ with a homogeneous point $\bar{\bfz} = (\bar{u},\bar{v},\bar{w})^T$ in any of the image planes $\bar{\bbP}_{\l}$ and $\bar{\bbP}_{\r}$ through a matrix-vector product:
\eqnl{eq:matrix_vector_P}{
\bar{\bfz} \propto P\bar{\bfx},
}
where $P$ is a $3 \times 4$ matrix and where ``$\propto$'' refers to equality up to a scaling factor. Homogeneous coordinates simplify the notation needed to describe perspective projections and allow for projective-geometric concepts such as points and lines at infinity \cite{Hartley2003}. For the purposes of Bayesian estimation, however, the perspective projection must be expressed in Euclidean coordinates, in order to allow for a meaningful definition of a distance between points, namely the Euclidean distance. The point $\bar{\bfz}$ is then expressed in Euclidean coordinates as $\bfz=(u,v)^T = (\bar{u}/\bar{w},\bar{v}/\bar{w})^T$ which is thus a nonlinear function of the coordinates of the real-world point $\bfx$. If $P$ is the projection onto the left (resp.\ right) image plane, then $\bfz$ will be the point $\bfz_{\l}$ (resp.\ $\bfz_{\r}$).

\section{Disparity Space}
\label{sec:disparity_space}

\begin{figure}
\footnotesize
\def\svgwidth{0.97\columnwidth}
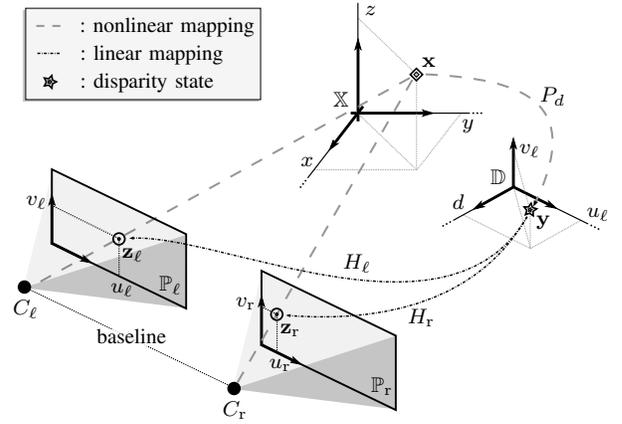
\caption{A rectified camera pair $(C_{\l},C_{\r})$ and the related disparity space $\bbD$.}
\label{fig:disparity_space}
\end{figure}

The concept of disparity space is closely linked to the idea of a rectified camera setup, as exemplified in Figure \ref{fig:disparity_space} ({\it cf.} Figure \ref{fig:triangulation}, showing a more general, non-rectified camera setup). Formally, assuming that the projection matrix $P_{\l}$ of the left camera $C_{\l}$ is of the form $P_{\l} = K[I\;\bfO]$, then the pair $(C_{\l},C_{\r})$ is called horizontally (resp.\ vertically) rectified if the projection matrix $P_{\r}$ of the right camera $C_{\r}$ is of the form $P_{\r} = K[I\;\bft]$ where $\bft = (b,0,0)^T$ (resp.\ $\bft = (0,b,0)^T$); the parameter $b$ is called the baseline. Henceforth, we will consider rectified cameras to be horizontally rectified, as in Figure \ref{fig:disparity_space}. Let $(C_{\l},C_{\r})$ be the rectified camera pair, let $\bbP_{\l}$ and $\bbP_{\r}$ be the respective camera image planes, and let the projections of a real-world point $\bfx$ in $\bbX$ be denoted with $\bfz_{\l}=(u_{\l},v_{\l})^T$ in $\bbP_{\l}$ and $\bfz_{\r}=(u_{\r},v_{\r})^T$ in $\bbP_{\r}$. The point $\bfx$ is represented in the disparity space $\bbD = \bbR^3$ associated to the rectified camera pair $(C_{\l},C_{\r})$ by a point $\bfy$ of the form
\eqns{
\bfy = \left(u_{\l},v_{\l}, d \right)^T,
}
where $d = u_{\r} - u_{\l}$ is referred to as the {\it disparity}, as it measures the difference in the camera views of the point $\bfx$. The point $\bfy$ characterises both the left and right projections $\bfz_{\l}$ and $\bfz_{\r}$, as depicted in Figure \ref{fig:disparity_space}.

In the context of projective geometry, it is possible to relate the points $\bfx$ and $\bfy$ through a linear transformation $P_{\d}$ as
\eqnl{eq:3d_to_disp}{
\bar{\bfy} \propto P_{\d} \bar{\bfx},
}
where $\bar{\bfy}$ and $\bar{\bfx}$ denote, as in the previous section, the projective equivalents of the points $\bfy$ and $\bfx$.

It is useful to express the transformation $P_{\d}$ in terms of the elements of the camera projection matrices $P_{\r}$ and $P_{\l}$. As a consequence of the fact that the camera pair is horizontally rectified, it holds that
\eqns{
(P_{\l})_{i\cdot} = (P_{\r})_{i\cdot},
}
for $i = 2,3$, where $(P)_{i\cdot}$ is the $i$\textsuperscript{th} row of the matrix $P$. The matrix $P_{\d}$ can then be expressed as
\eqnl{eq:transformation_pd}{
P_{\d} =
\begin{bmatrix}
(P_{\l})_{1\cdot} \\
(P_{\l})_{2\cdot} \\
(P_{\r})_{1\cdot} - (P_{\l})_{1\cdot} \\
(P_{\l})_{3\cdot}
\end{bmatrix}.
}

The existence of transformation $P_{\d}$ means that the disparity space $\bbD$ can be used as a proxy space for triangulation from cameras and any point in $\bbD$ can be converted to its equivalent in $\bbX$ via the inverse transform of $P_{\d}$.

To allow for triangulation, a link between the disparity space and the image planes must also be established. With the rectified camera setup, the point $\bfy$ is projected onto the left- and right-camera image plane, $\bbP_{\l}$ and $\bbP_{\r}$, by applying the respective orthographic projections, $H_{\l}$ and $H_{\r}$, defined as
\eqnl{eq:linear_mapping}{
H_{\l} =
\begin{bmatrix}
1 & 0 & 0\\
0 & 1 & 0
\end{bmatrix},
\quad\mbox{ and }\quad
H_{\r} =
\begin{bmatrix}
1 & 0 & 1\\
0 & 1 & 0
\end{bmatrix}.
}

In summary, the disparity space associated with a pair of rectified cameras allows for expressing the process of observation (\ref{eq:matrix_vector_P}) as a linear mapping (\ref{eq:linear_mapping}), while maintaining a one-to-one correspondence with the real world $\bbX$, as shown in Equation (\ref{eq:3d_to_disp}).

\section{Modelling uncertainty}
\label{sec:modelling_uncertainty}

The purpose of this section is to describe the sources of uncertainty in an object state estimate in 3D estimation from cameras.
The state of the object of interest is composed of intrinsic and/or extrinsic parameters which characterise the object and its behaviour.



\subsection{Static object}
\label{ssec:static_object}

The most common approach in Bayesian tracking is to assume that objects are point-like. This is justified in radar applications by the relatively small extent of the objects in the scene, when compared to the radar resolution. In such a case, if an object of interest is static, its state can be described by its position, represented by a point state in $\bbX$ \cite{RisticEtAl2004}.

When the sensor is a camera, the extent of the objects is often observable, so that the shape of the objects can be estimated \cite{Koch2008}. Yet, in the context of camera calibration, the estimation of the extent, or of the shape, of the objects is not always desirable, as it significantly increases the difficulty of the problem without directly contributing to the convergence of the estimation of calibration parameters. We approach this problem by modelling the extended observation of an object on the camera image planes as an uncertainty on the point-like object state, which affects the estimation in a similar way to a point spread function. As a consequence, we cannot use a point in $\bbX$ to describe the object, but have to resort to a probability distribution $p$ on $\bbX$.
 
Another source of uncertainty in the point state of the object stems from the observation process itself, namely the camera observations, which are known to be noisy due to the nature of the imaging process.

\begin{figure}
\centering
\includegraphics[width=0.97\columnwidth]{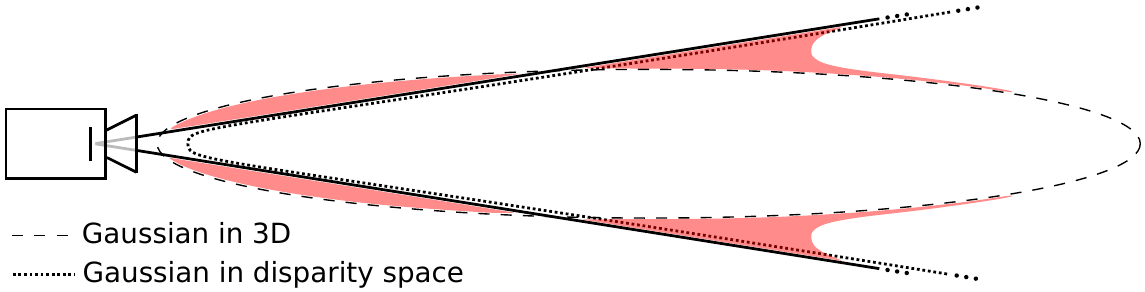}
\caption{Modelling of uncertainty in triangulated camera observations. The solid lines show the actual uncertainty in 3D, defined by the camera's field of view. When the Gaussian uncertainty in camera observations is mapped from the image space, via disparity space~(\ref{eq:linear_mapping}), into the 3D space $\bbX$~(\ref{eq:transformation_pd}), it takes on a distinctly non-Gaussian nature, as shown by the dotted curve. In contrast, the corresponding Gaussian uncertainty in 3D is shown by the dashed ellipse and the difference between the Gaussian and non-Gaussian representation is highlighted in red.}
\label{fig:GaussianRep}
\end{figure}

We model observation errors in $\bbP_{\l}$ and $\bbP_{\r}$ as Gaussian, which is generally a reasonable assumption~\cite{SibleyEtAl2007}. It follows that the corresponding uncertainty in the space $\bbX$ is non-Gaussian, as illustrated in Figure~\ref{fig:GaussianRep}.
This raises the question of how to accurately characterise the distribution $p$ on $\bbX$. Although the Gaussian distribution is very popular, it is clearly not appropriate in this context. The choice of a good model for $p$ is further complicated by the fact that the uncertainty in the triangulated object state is range-dependent, {\it i.e.}, heteroscedastic. In such circumstances, one typically resorts to particle representations~\cite{Arulampalam2002} to approximate $p$.

However, the particle representation of $p$ also has its limitations. One of the most serious limitations is its inability to represent objects that are infinitely far away from the camera. In this case, the support of the distribution $p$ is not bounded and infinitely many particles are required to represent it fairly.

The inapplicability of the usual representations to modelling the distribution $p$ in $\bbX$ motivates the use of {\it another state space}, the {\it disparity space}, in which this can be achieved more easily.

As shown in the previous section, the disparity space $\bbD$ is related to the camera image planes $\bbP_{\l}$ and $\bbP_{\r}$ via a linear transformation~(\ref{eq:linear_mapping}). It follows directly that a Gaussian uncertainty in these image planes back-transforms into a Gaussian distribution on $\bbD$. The fact that $\bbD$ is also in one-to-one relation with $\bbX$ through Equation (\ref{eq:3d_to_disp}) makes the disparity space a suitable space for the representation of the uncertainty for purposes of 3-D estimation. Estimating the position of the object of interest can then be achieved via a Kalman filter update, as demonstrated in~\cite{Clark2010} and~\cite{Ivekovic2009}.

As the concept of disparity is related to the concept of inverse depth, the disparity space $\bbD$ inherits from the advantages of any of the inverse-depth based parametrisations \cite{Civera2008,Montiel2006}, but it also enables a linear projection onto the image planes $\bbP_{\l}$ and $\bbP_{\r}$, making it particularly suitable for Bayesian tracking.

\subsection{Dynamic object}
\label{ssec:dynamic_object}

\begin{figure}
\centering
\subfigure[$u$-$v$ plane in the disparity space $\bbD$]{\label{fig:EKF_UKF_particle_UV}
\includegraphics[trim=125pt 270pt 115pt 270pt,clip,width=0.8\columnwidth]{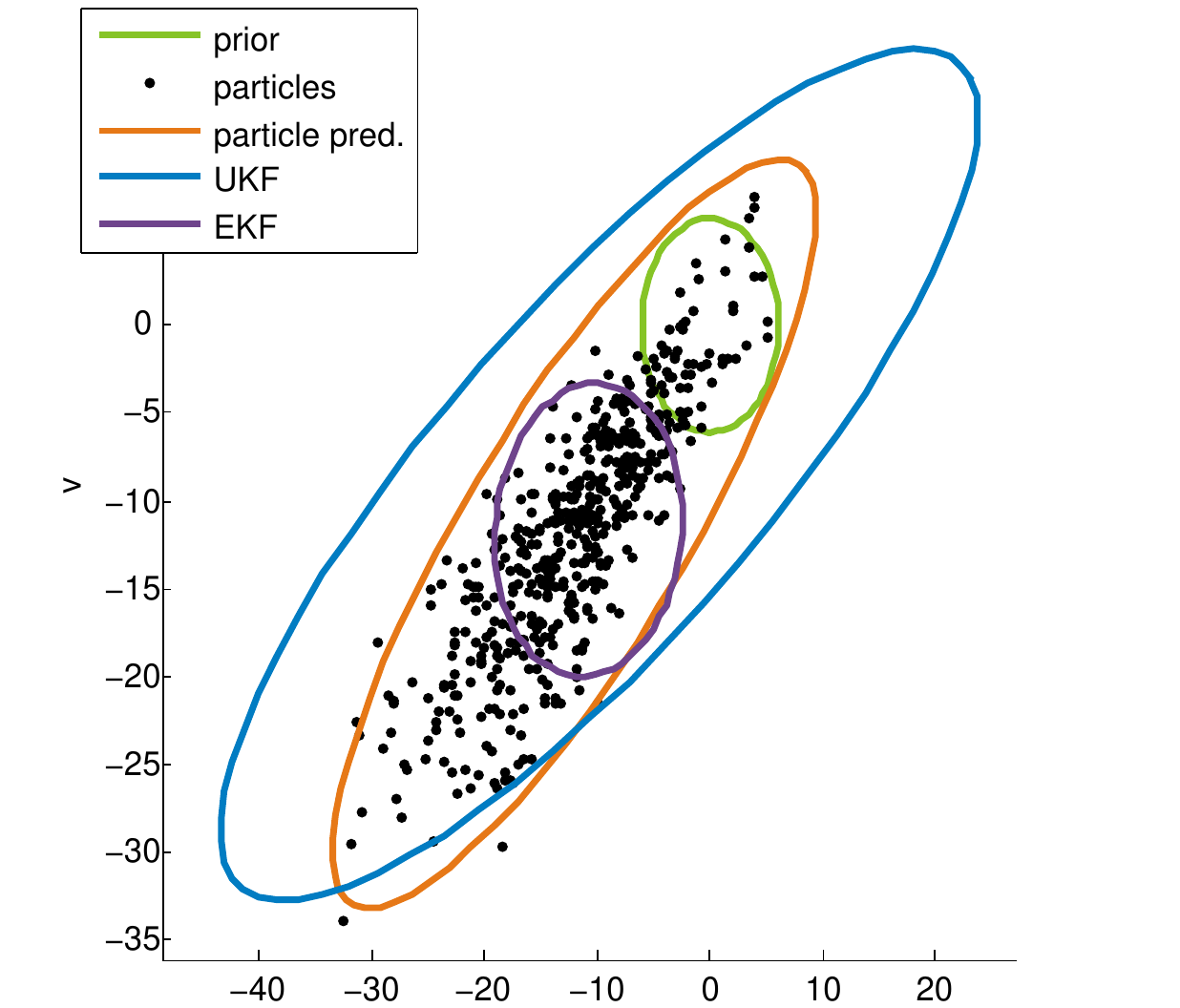}}
\subfigure[$v$-$d$ plane in the disparity space $\bbD$]{\label{fig:EKF_UKF_particle_VD}
\includegraphics[trim=125pt 255pt 115pt 270pt,clip,width=0.8\columnwidth]{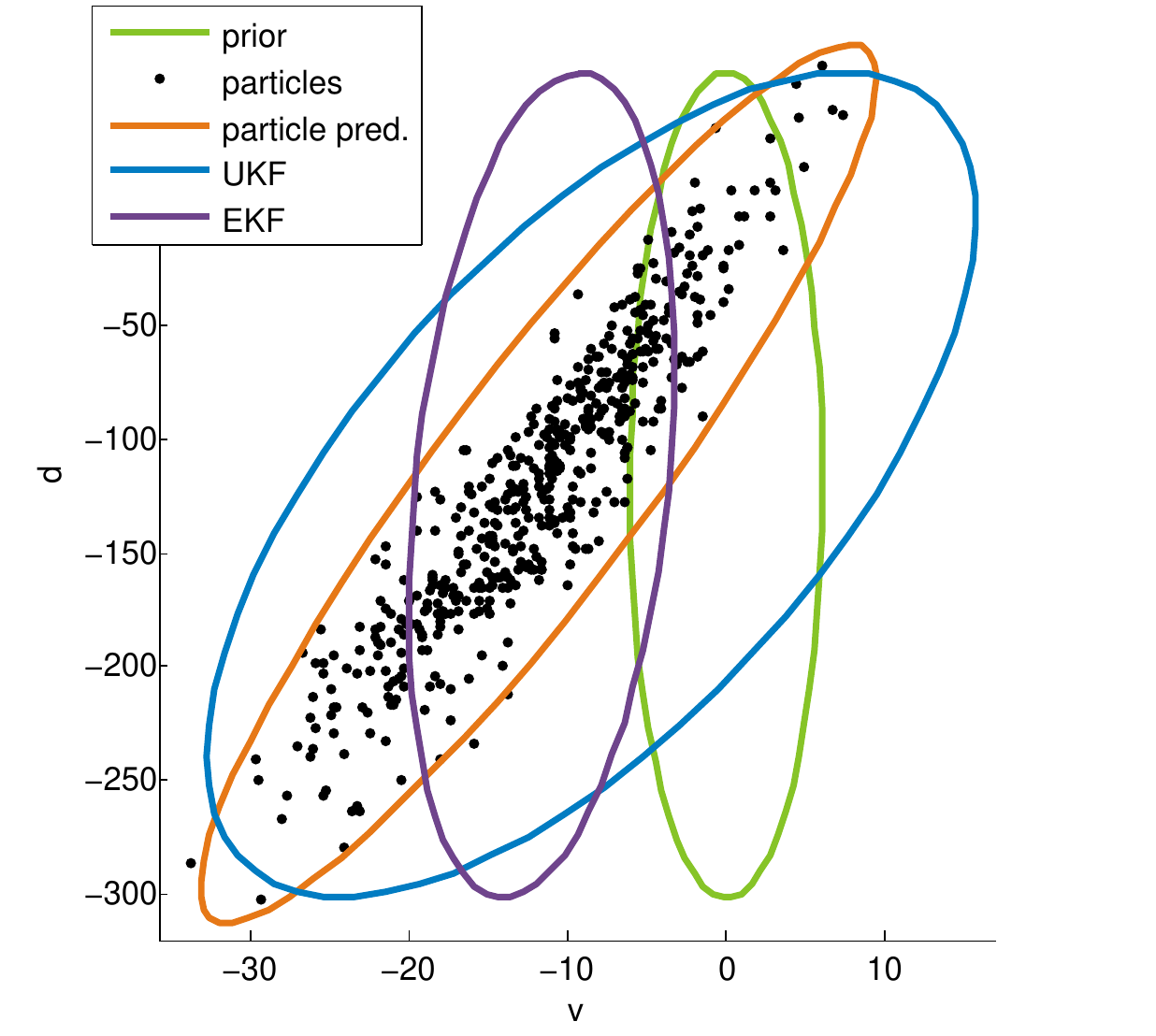}}
\caption{Transformation of a Gaussian prior (green) for the prediction of a dynamical object. The model is constant velocity with $\dot{x} = \dot{y} = 2\,\cm.\s^{-1}$ and $\dot{z} = 0.5\,\cm.\s^{-1}$ and a noise with variance $0.08\,\cm^2.\s^{-2}$.}
\label{fig:EKF_UKF_particle}
\end{figure}

If the object of interest is dynamic, its state will include its position in $\bbX$, as well as parameters modelling its dynamics.

The dimensionality of the state space depends on the type of dynamics required to model the motion of the object. For instance, if the object is assumed to have a constant, but unknown, velocity, then the state of the object is a vector in $\bbR^6$, where the $2\times3$ coordinates represent the object's position and velocity in $\bbX$. Therefore, let $\hat{\bbX} = \bbR^6$ be the space of vectors of the form $\bfx = (x,y,z,\dot{x},\dot{y},\dot{z})^T$, where $\dot{x}=dx/dt$, $\dot{y}=dy/dt$ and $\dot{z}=dz/dt$. The disparity space $\bbD$ also has to be extended to model velocity, and we define $\hat{\bbD} = \bbR^6$ as the space of vectors of the form $\bfy = (u_{\l},v_{\l},d,\dot{u}_{\l},\dot{v}_{\l},\dot{d})$, with $\dot{u}_{\l},\dot{v}_{\l}$ and $\dot{d}$ similarly defined as time derivatives of $u_{\l},v_{\l}$ and $d$.

Let $\bbT = \bbN$ be the set of time steps. The dynamics of the object of interest are usually uncertain and are modelled by a Markov transition $M_{t+1|t}: \hat{\bbX} \rightarrow \hat{\bbX}$, such that if the object is at point $\bfx \in \hat{\bbX}$ at time $t \in \bbT$, then the probability for it to be at point $\bfx'$ at time $t+1$ is $M_{t+1|t}(\bfx'|\bfx)$.

The uncertainty associated with the object's dynamic transition from $t$ to $t+1$ is often assumed to be Gaussian in $\hat{\bbX}$. As discussed in Section \ref{ssec:static_object}, however, the uncertainty in the position of the object is more naturally represented as a Gaussian in $\bbD$. This raises the following question: how to relate these two types of uncertainty in the estimation process?

Denoting with $p_t$ the Gaussian distribution on $\hat{\bbD}$, representing the state of the object at time $t$, the proposed solution to this question can be decomposed into the following 6 steps:\\
\begin{enumerate*}[label=\itshape\alph*\upshape)]
\item \noindent Sample a particle representation of $p_t$ in $\hat{\bbD}$;\\
\item Map this representation into $\hat{\bbX}$;\\
\item Apply the Markov transition $M_{t+1|t}$ in $\hat{\bbX}$;\\
\item Map the resulting particle representation back into $\hat{\bbD}$;\\
\item \label{step:sampleGauss} Recover the Gaussian distribution $p_{t+1|t}$ by computing the mean and covariance of the resulting representation in $\hat{\bbD}$;\\
\item Compute $p_{t+1}$ by applying the Kalman update in $\hat{\bbD}$.\\
\end{enumerate*}

Although the approach described in Algorithm \ref{algo:single_rectified} bears some similarity with the UKF, in the sense that we are approximating a Gaussian with particles, the important difference is that samples are drawn randomly from the posterior, which enables us to maintain the nonlinearity when reparameterising. The reason for not using the UKF itself, or even an Extended Kalman Filter (EKF), is that the non-linearity in the observation model is too pronounced to be fairly represented by a point (EKF) or by a few $\sigma$-points (UKF). The approximation that is applied in step \ref{step:sampleGauss} above, by recovering a Gaussian distribution from the particle representation, might be very optimistic, yet the objective is to be explicitly aware of the uncertainty, which might not be the case with the UKF and EKF. This aspect is exemplified in Figure \ref{fig:EKF_UKF_particle}, where the distribution before and after prediction in the $u$-$v$ and $v$-$d$ planes is displayed for several prediction methods. In the case depicted, the EKF manages to capture the overall motion as the mean of the associated Gaussian distribution and the mean of the set of particles seem to match, yet, it fails to understand the evolution of the uncertainty and, e.g., clearly underestimate it in the $u$-$v$ plane. In the case of UKF, it appears that even though the shift of the mean and the general evolution of the uncertainty is better captured than with the EKF, there is still a non-negligible error in the estimation of the covariance. This is mainly due to the noise on the motion model in $\bbX$, which becomes non-linear in the disparity space $\bbD$. The particle prediction which relies on $500$ particles, manages to capture both the non-linearity of the motion and of the associated noise.

\section{Single-object estimation}
\label{sec:single_object_estimation}

\subsection{Rectified camera pair}

As mentioned in the previous section, the estimation of a single static object can be handled via a Kalman filter update in $\bbD$. A \emph{particle prediction} between two time steps is used when the object is dynamic ({\it i.e.}, when its state lives in $\hat{\bbD}$). Let $(\bfz_i,R_i)$ be the mean and covariance of an observation at time $t \in \bbT$ from the camera $C_i$, with $i \in \{\l,\r\}$. The likelihood $L^i_t(\bfz_i|\bfy)$ can be expressed as
\eqnsa{
L^i_t(\bfz_i|\bfy) = \calN(\bfz_i; H_i\bfy, R_i),
}
where $\calN(\bfx;\bfm,P)$ is a normal distribution with mean $\bfm$ and covariance matrix $P$, evaluated at point $\bfx$. If the object is dynamic, the observation matrices~(\ref{eq:linear_mapping}) have to be suitably augmented. For example,\
\eqns{
H_{\l} =
\begin{bmatrix}
1 & 0 & 0 & 0 & 0 & 0\\
0 & 1 & 0 & 0 & 0 & 0
\end{bmatrix},
}
is the observation matrix from $\hat{\bbD}$ to the left camera image plane $\bbP_{\l}$. Note that the velocity is assumed to be unobserved.

In order to completely specify an estimation algorithm for the object of interest, the initialisation has to be described as well. As we do not assume that the camera pair $(C_{\l},C_{\r})$ is synchronised, initialisation has to be dealt with using a single camera, say the left camera $C_{\l}$. Consider that we receive the first observation $\bfz_{\l} = (u_{\l},v_{\l})$ with covariance $R_{\l}$ at time $t = 0$. This observation can be used directly to initialise the first two components of the Gaussian distribution $p_0$ in $\bbD$. However, the disparity, and possibly the velocity, are not known a priori and have to be initialised in some other way. The mean of the disparity can be computed by considering the expected distance between the left camera $C_{\l}$ and the object. The variance has to be taken sufficiently large for the disparity $0$ to be likely enough, whenever the object is possibly infinitely far away from the camera. As a consequence, negative disparity, which represents objects behind the camera pair, must be included. This is necessary in order to maintain a Gaussian distribution in $\bbD$ and does not represent an issue in general. The mean and covariance $\bfy_0$ and $Q_0$ of the Gaussian distribution $p_0$ can now be determined, and the estimation carried out, as described in Algorithm \ref{algo:single_rectified}.

\begin{algorithm}
\KwData{
\begin{itemize}
\item Gaussian distribution $(\bfy_{t-1},Q_{t-1})$,
\item Observation $\bfz_i \in \bbP_i$ with uncertainty $R_i$, where $i$ is $\l$ or $\r$.
\end{itemize}
}
\KwResult{Initialised/updated Gaussian distribution at time $t$}
\eIf{$t==0$}{
$[(\bfy_0,Q_0)]$ = \texttt{initialisation}$[(\bfz_i,R_i)]$
}{
	\eIf{object is static}{
		$(\hat{\bfy}_t,\hat{Q}_t)$ = $(\bfy_{t-1},Q_{t-1})$;
	}{
		$[(\hat{\bfy}_t,\hat{Q}_t)]$ = \texttt{particle\_prediction}$[(\bfy_{t-1},Q_{t-1})]$
	}
$[(\bfy_t,Q_t)]$ = \texttt{Kalman\_update}$[(\hat{\bfy}_t,\hat{Q}_t),(\bfz_i,R_i)]$
}
\caption{Single-object estimation with a rectified camera pair at time $t \in \bbT$.}
\label{algo:single_rectified}
\end{algorithm}

\subsection{Non-rectified camera pair}

\begin{figure}
\centering
  \subfigure[$b = 0.5m$.]{
    \includegraphics[trim=125pt 270pt 120pt 285pt,clip,width=0.45\columnwidth]{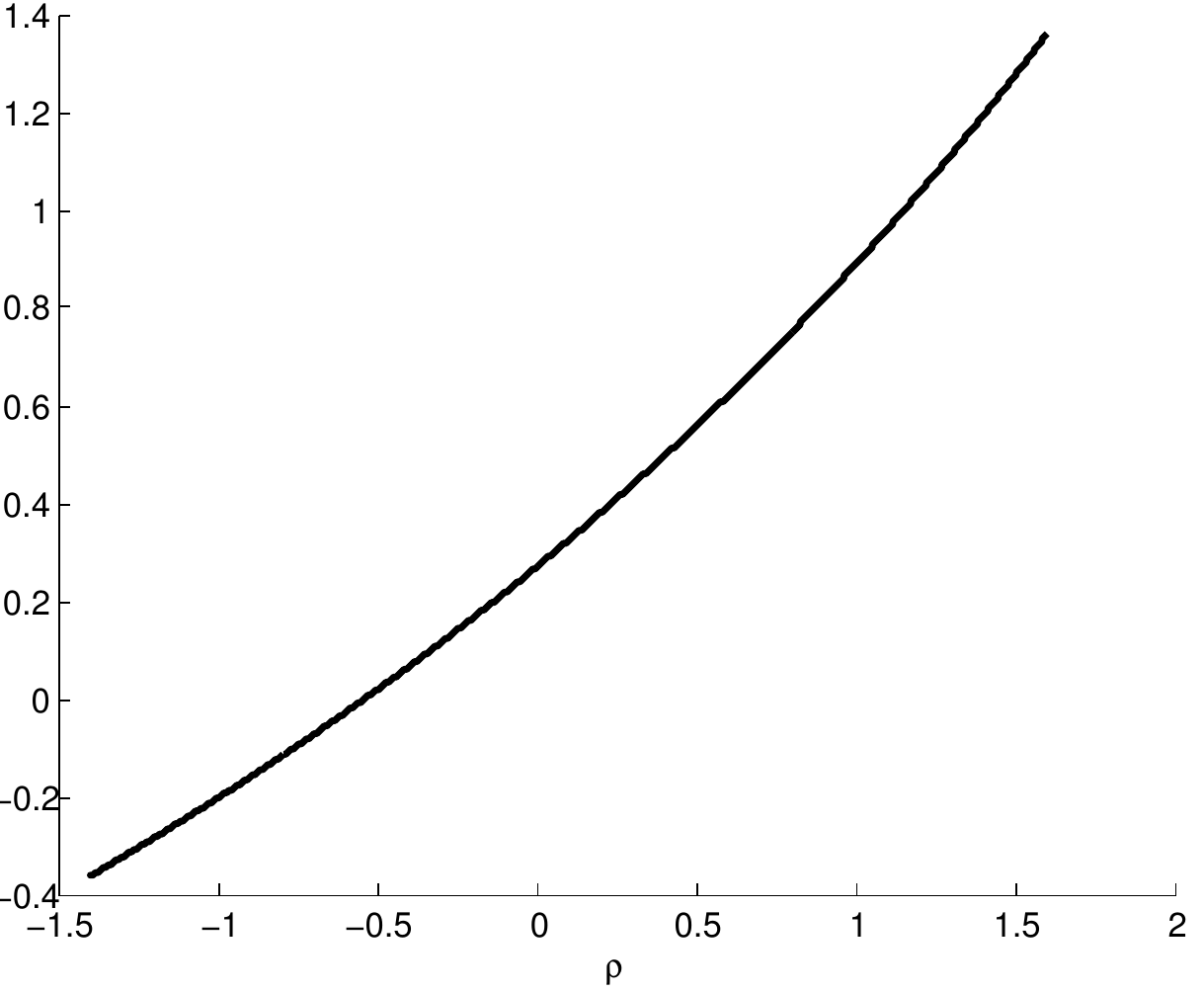}
  }
  \subfigure[$b = 5m$.]{
    \includegraphics[trim=125pt 270pt 120pt 285pt,clip,width=0.45\columnwidth]{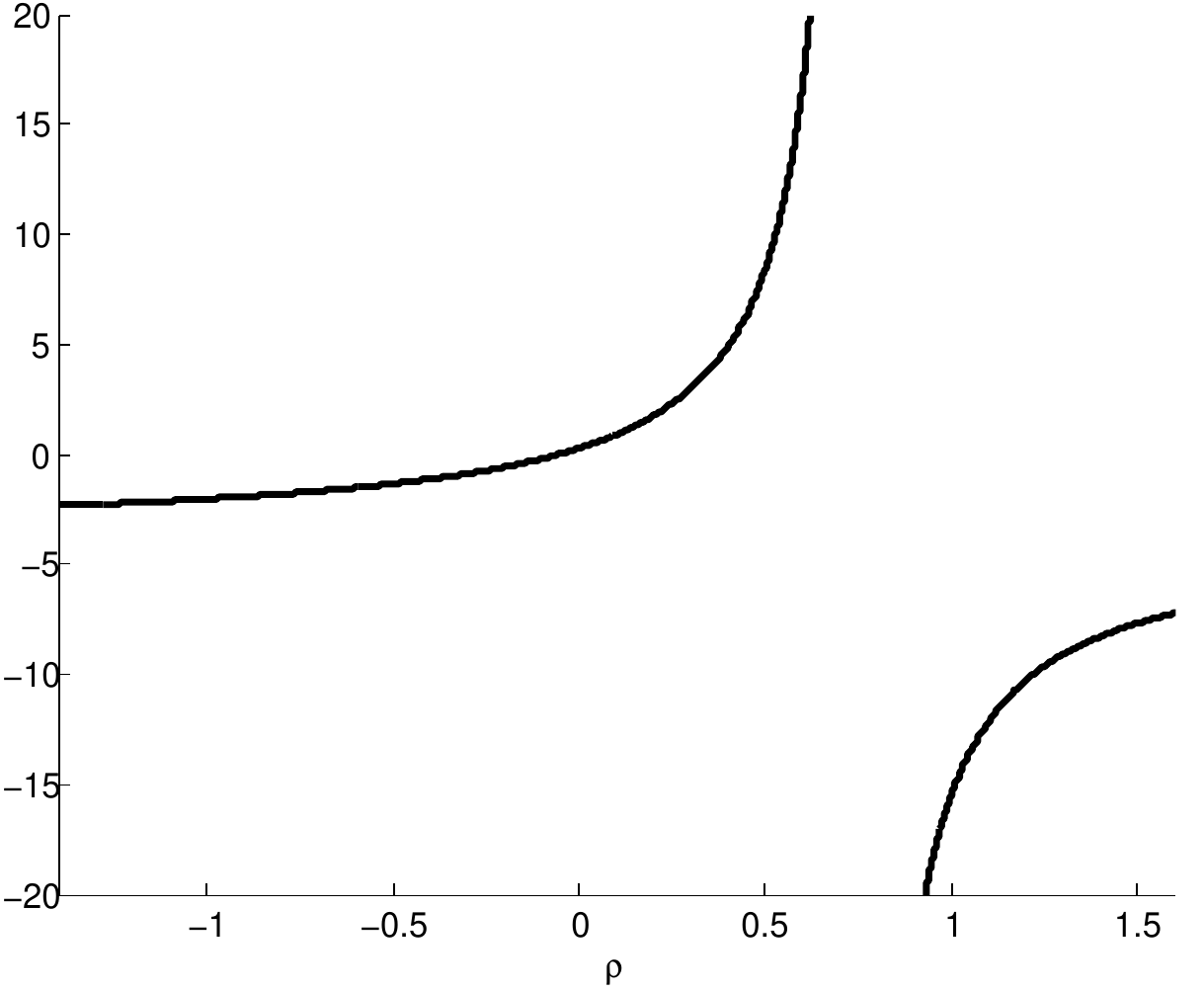}
  }
\caption{Illustration of the non-linearity of the observation as a function of the inverse-depth $\rho$, as the camera pair becomes non-rectified. Starting from a rectified configuration with baseline $b$, the right camera $C_{\r}$ is rotated by an angle of $\pi/12$ radians.}
\label{fig:nonLinearity}
\end{figure}

Estimating the state of an object from a non-rectified camera pair $(C_{\l},C_{\r})$ is a challenging problem, as the linear observation model obtained from the rectified camera geometry is not available anymore. This aspect is illustrated in Figure \ref{fig:nonLinearity}, where the nonlinearity of the observation function is shown for two different non-rectified camera pairs. Yet, taking advantage of the approach which applies in the rectified case, and has been detailed in the previous sections, is still beneficial. This idea is described in detail and assessed against the standard inverse-depth parametrisation in \cite{Houssineau2012}, so that only the underlying principles are restated here.

In the previous section, a particle-based prediction has been used in order to handle the possible motion of the object of interest. In the case of a non-rectified camera pair, a similar idea can be used to map the distribution from a disparity space specifically constructed for the left camera $C_{\l}$ to another disparity space, constructed for the right camera $C_{\r}$.

\begin{figure}
\centering
\footnotesize
\def\svgwidth{0.97\columnwidth}
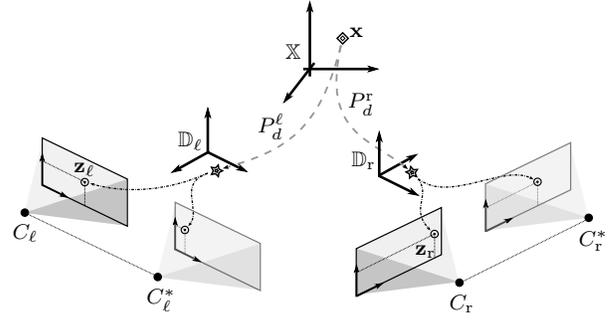
\caption{Projection of a point $\bfx$ in $\bbX$ onto the disparity spaces $\bbD_{\l}$ and $\bbD_{\r}$ and onto the image planes of the two rectified camera pairs $(C_{\l},C^*_{\l})$ and $(C_{\r},C^*_{\r})$.}
\label{fig:nonrectified_triangulation}
\end{figure}

The properties of disparity spaces are still strong assets, even when considering a single camera. Yet, a disparity space requires two cameras in order to be defined. The idea is then to introduce two \emph{abstract} cameras $C^*_{\l}$ and $C^*_{\r}$ that are rectified with respect to the left and right cameras, respectively. These cameras are said to be abstract as they do not exist physically, and hence never produce observations. Two disparity spaces $\bbD_{\l}$ and $\bbD_{\r}$ are thus defined based on the rectified camera pairs $(C_{\l},C^*_{\l})$ and $(C_{\r},C^*_{\r})$ and are related to $\bbX$ via the projective transformations $P_{\d}^{\l}$ and $P_{\d}^{\r}$, as shown in Figure \ref{fig:nonrectified_triangulation}. The process of predicting a probability distribution while starting from the disparity space $\bbD_{\l}$ (resp.\ $\bbD_{\r}$) and arriving into the disparity space $\bbD_{\r}$ (resp.\ $\bbD_{\l}$) will be called a \emph{particle move}. Indeed, the principle of this approach is to use particle representations in order to perform the mapping of the Gaussian distribution representing the object of interest from one disparity space to another.

As mentioned before, this approach for a static object has been assessed in \cite{Houssineau2012}. However, its use for a moving object, as in Algorithm \ref{algo:single_nonrectified}, is novel. The performance of this extension will be evaluated, together with other generalisations, in Section \ref{sec:simulated_data}.

\begin{algorithm}
\KwData{
\begin{itemize}
\item Gaussian distribution $(\bfy^j_{t-1},Q^j_{t-1})$ in $\bbD_j$, $j \in \{\l,\r\}$,
\item Observation $\bfz_i \in \bbP_i$ with uncertainty $R_i$, $i \in \{\l,\r\}$.
\end{itemize}
}
\KwResult{Initialised/updated Gaussian distribution at time $t$}
\eIf{$t==0$}{
$[(\bfy^i_0,Q^i_0)]$ = \texttt{initialisation}$[(\bfz_i,R_i)]$
}{
	\eIf{i == j}{
		$[(\hat{\bfy}^i_t,\hat{Q}^i_t)]$ = \texttt{particle\_prediction}$[(\bfy^j_{t-1},Q^j_{t-1})]$
	}{
		$[(\hat{\bfy}^i_t,\hat{Q}^i_t)]$ = \texttt{particle\_move}$[(\bfy^j_{t-1},Q^j_{t-1})]$
	}
$[(\bfy^i_t,Q^i_t)]$ = \texttt{Kalman\_update}$[(\hat{\bfy}^i_t,\hat{Q}^i_t),(\bfz_i,R_i)]$
}
\caption{Single-object estimation with a non-rectified camera pair at time $t \in \bbT$.}
\label{algo:single_nonrectified}
\end{algorithm}

\begin{figure}
\centering
\subfigure[$u$-$v$ plane in the disparity space $\bbD_{\l}$]{\label{fig:UKF_particle_UV}
\includegraphics[trim=140pt 255pt 115pt 270pt,clip,width=0.8\columnwidth]{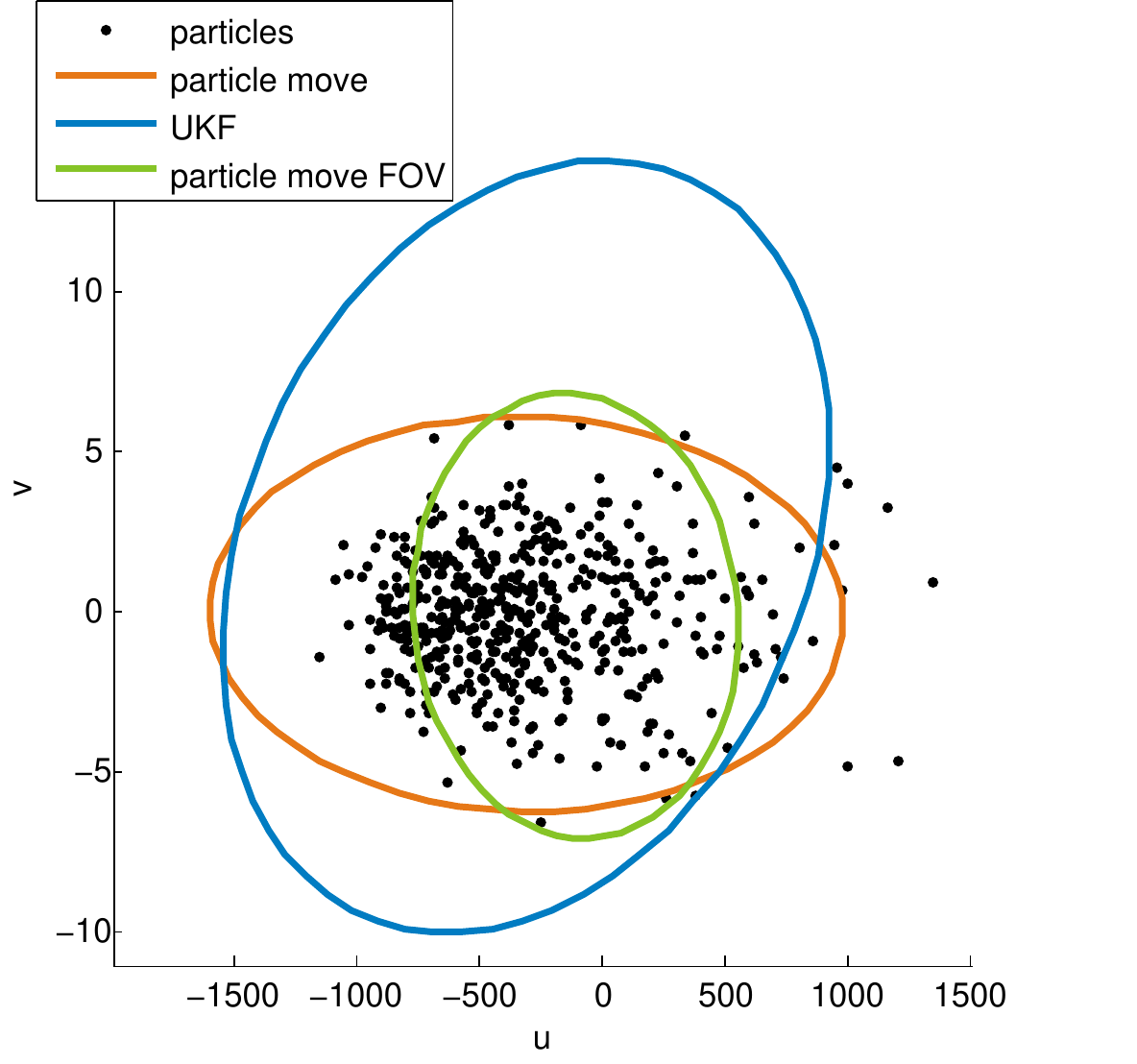}}
\subfigure[$v$-$d$ plane in the disparity space $\bbD_{\l}$]{\label{fig:UKF_particle_VD}
\includegraphics[trim=135pt 255pt 115pt 270pt,clip,width=0.8\columnwidth]{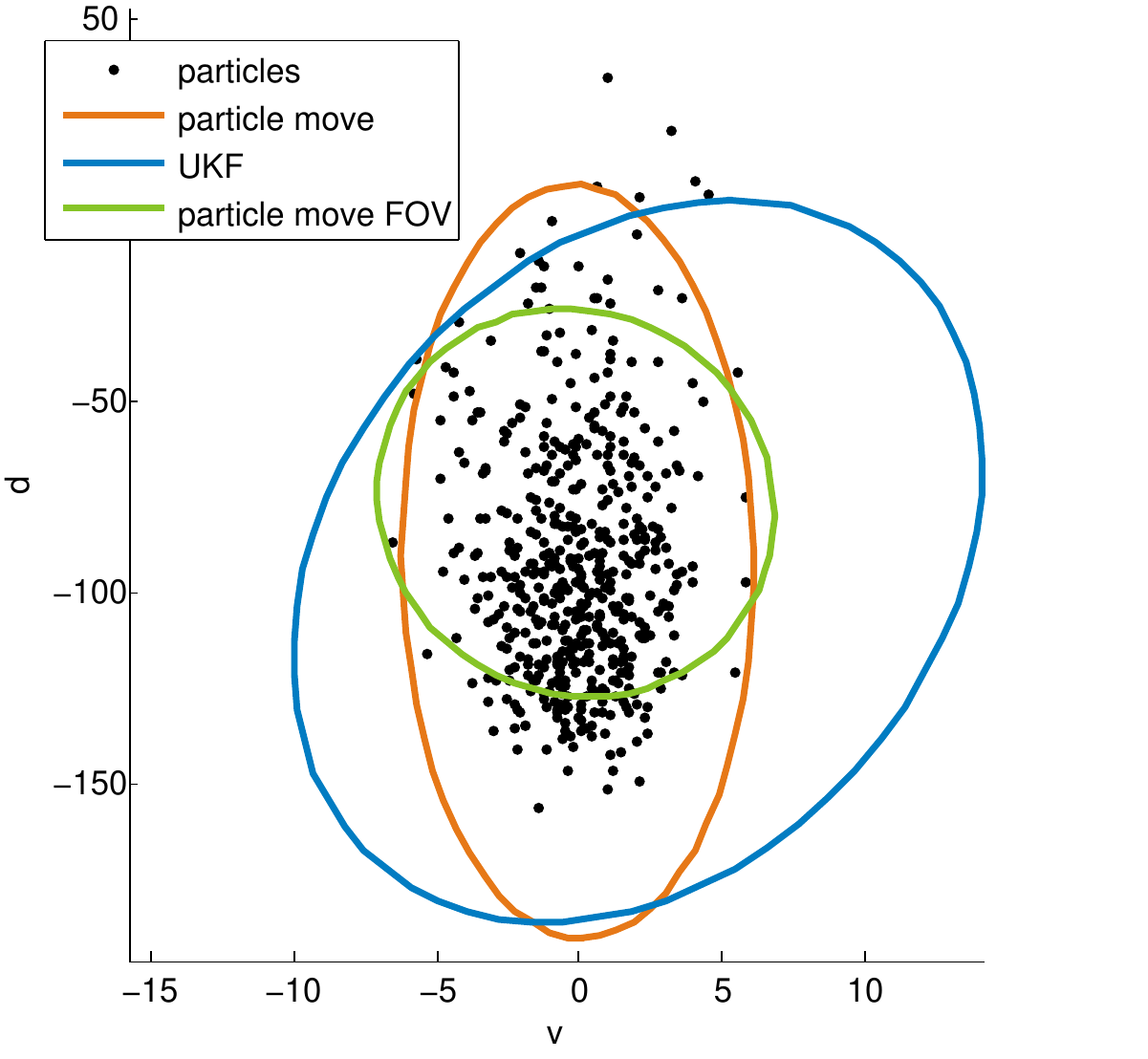}}
\caption{Transformation of a Gaussian prior in the Disparity space $\bbD_{\r}$ to the disparity space $\bbD_{\r}$ for the prediction of a dynamical object. The two cameras have the same intrinsic parameters, with a resolution of $800\times600$ and a focal length of $8\mm$. The distance between the two cameras is $200\,\cm$ and the yaw is $-\pi/8$ for the left camera and $\pi/8$ for the right camera. The velocity has mean $0$ and variance $0.03\,\cm^2.\s^{-2}$.}
\label{fig:UKF_particle}
\end{figure}

Figure \ref{fig:UKF_particle} illustrates the form of the distributions obtained when handling the mapping from $\bbD_{\l}$ to $\bbD_{\r}$ with different methods. $500$ particles have been used to represent the actual distribution in $\bbD_{\r}$. Once again, the UKF shows inaccuracies in its representation of the objective distribution even though the noise on the motion is lower than for the example shown in Figure \ref{fig:EKF_UKF_particle}. This can be explained by the non-linearity of the mapping between $\bbD_{\l}$ and $\bbD_{\r}$, which makes the representation of the uncertainty even more difficult. Note that, in Figure \ref{fig:UKF_particle_UV}, the range on the axis $u$ is much larger than on the axis $v$, and actually extends outside of the field of view of the right camera. When estimating a single object, we assume that the object is detected and hence the particles outside of the field of view can be discarded before fitting a Gaussian distribution, as represented by the green ellipse in Figure \ref{fig:UKF_particle}. This approach is described in Section \ref{ssec:state_dependent_detection}.

Although we have described the procedure for two cameras, the approach can be straightforwardly extended to more cameras by introducing a disparity space for each camera.

\section{Multi-object estimation}
\label{sec:multi_object_estimation}

The previous sections have described the disparity space parametrisation and how to use this for statistically triangulating and tracking a single object. In this section, the approach is extended to account for multiple objects in the scene, where there may be false alarms and missed detections from the sensor. This will underpin the method for camera calibration in the next section, since it provides the likelihood to update the probability density on the sensor parameters.

Throughout this section, objects will be considered to be possibly moving according to a constant velocity model, so that the spaces $\bbD_{\l}$ and $\bbD_{\r}$ are augmented with velocity and respectively denoted with $\hat{\bbD}_{\l}$ and $\hat{\bbD}_{\r}$.

\subsection{General solution}
\label{ssec:general_solution}

We consider a population $\calX_t$, defined as the set of objects of interest in the scene, at time $t \in \bbT$. Most often, the size of the population $\calX_t$ is not known and might vary in time. Additionally, the correspondences between the estimated population and the received observations are not generally known. As a consequence, a sufficiently general model has to be constructed in order to allow for the estimation of the population $\calX_t$ for any time $t \in \bbT$. 

The most popular estimation framework applicable in this context is the FInite Set STatistics (FISST)~\cite{Mahler2007}. In the following, we provide only a brief summary of the FISST framework, necessary to motivate the remainder of the paper, and refer the interested reader to~\cite{Houssineau2013,Mahler2007} for a more exhaustive description.

 Within FISST, it is possible to model that:
\begin{enumerate}
\item a new set of objects $\calX^{\ib}_t$ might appear a each time $t \in \bbT$, so that $\calX_t = \calX_{t-1} \cup \calX^{\ib}_t$,
\item every object's motion is independent of the other objects,
\item an object in $\calX_{t-1}$ with state $\bfy' \in \hat{\bbD}_i$ might disappear from the scene with probability $p_S(\bfy')$,
\item an object in $\calX_t$ with state $\bfy \in \hat{\bbD}_i$ can be either non detected with probability $1-p^i_D(\bfy)$ or detected through the observation $\bfz \in \bbP_i$ with probability $p^i_D(\bfy)L^i_t(\bfz|\bfy)$, with $i \in \{\l,\r\}$, and,
\item the set $Z^i_t$ of observations in $\bbP_i$ at time $t$ contains independent object-originated observations, as well as independent spurious observations, spatially distributed according to the probability density $c_i$ on $\bbP_i$, and the number of which is driven by a Poisson distribution with parameter $\lambda_i$.
\end{enumerate}
We assume that $p^i_D$ only depends on the coordinates $(u_i,v_i)$ in the image plane $\bbP_i$, so that the choice of state space has no consequence on the probability of detection.

As the correspondences between objects and observations are not assumed to be known, we introduce association functions $\theta : Y \to \phi \cup Z^i_t$, where $\phi$ is the empty observation. Denoting with $Y_z$ the inverse image of $Z^i_t$ through $\theta$, we assume that the restriction $\theta|_{Y_z} : Y_z \to Z^i_t$ of the function $\theta$ is a bijection. The set of such association functions is denoted with $\Theta$.

With these models and assumptions, and following \cite{Mahler2007}, we can proceed to the estimation of the population via the following prediction and update steps:
\eqnsa{
\hat{P}^i_t(Y) & = \int \bfM^{i|j}_{t|t-1}(Y|Y') P^j_{t-1}(Y') \delta Y',\\
P^i_t(Y) & = \dfrac{\bfL^i_t(Z_t|Y)\hat{P}^i_t(Y)}{\displaystyle \int \bfL^i_t(Z_t|Y') \hat{P}^i_t(Y') \delta Y'},
}
where $\int \cdot \delta Y$ refers to the set integral \cite{Mahler2007}, $\hat{P}^i_t(Y)$ and $P^i_t(Y)$ are the predicted and updated multi-object densities describing the probability for the objects in $\calX_t$ to be at given points in the set $Y$ of points in $\hat{\bbD}_i$, and $\bfM^{i|j}_{t|t-1}$ and $\bfL^i_t$ are the conditional multi-object densities, describing prediction and update, with $\bfL^i_t$ expressed as
\eqnsml{
\bfL^i_t(Z^i_t|Y) = e^{-\lambda_i} \bigg[\prod_{\bfz \in Z^i_t} \lambda_i c_i(\bfz) \bigg] \times \bigg[\prod_{\bfy \in Y} \left(1-p^i_D(\bfy)\right) \bigg]\\
\times \sum_{\theta \in \Theta} \bigg[ \prod_{\bfy \in Y_z}
\dfrac{p^i_D(\bfy) L^i_t(\theta(\bfy)|\bfy)}
{\left(1-p^i_D(\bfy)\right) \lambda_i c_i(\theta(\bfy))} \bigg].
}

The evaluation of every possible association in $\Theta$ is extremely costly in practice and the complexity becomes exponential in time. It is therefore useful to avoid resorting explicitly to $\Theta$. This is made possible by reducing the multi-object densities $\hat{P}^i_t$ and $P^i_t$ to their first moment densities $\hat{\mu}^i_t$ and $\mu^i_t$. With additional assumptions, the estimation can be performed using only these first moment densities, and the resulting filter is called the Probability Hypothesis Density filter, or PHD filter \cite{Mahler2003}.

\subsection{The PHD filter}
\label{ssec:phd_filter}

As stated in the previous section, it is possible, with some assumptions, to propagate only the first moment of the multi-object densities of interest. These assumptions are as follows.
\begin{enumerate}[label=\bfseries A.\arabic*]
\item At any time $t \in \bbT$, all the objects in $\calX_t$ have the same probability density $p^i_t$ on $\hat{\bbD}_i$, $i \in \{\l,\r\}$.
\item The cardinality distribution of the set $\calX_t$ follows a Poisson distribution.
\end{enumerate}

Under these two assumptions, and following \cite{Mahler2003}, the first-moment density describing the population of interest can be propagated as follows
\eqnsa{
\hat{\mu}^i_t(\bfy) & = \mu^{\ib}_t(\bfy) + \int p_S(\bfy') M^{i|j}_{t|t-1}(\bfy|\bfy') \mu^j_t(\bfy) \d \bfy, \\
\mu^i_t(\bfy) & = (1-p^i_D(\bfy)) \hat{\mu}^i_t(\bfy) \\
&\qquad+\sum_{\bfz \in Z^i_t} \int \dfrac{p^i_D(\bfy)L^i_t(\bfz|\bfy)\hat{\mu}^i_t(\bfy)}
{\lambda^i c^i(\bfz) + \int p^i_D(\bfy')L^i_t(\bfz|\bfy')\hat{\mu}^i_t(\bfy') \d\bfy'},
}
where $\mu^{\ib}_t$ is the first-moment density representing the appearing set of individuals $\calX^{\ib}_t$. 

Two implementations of the PHD filter are available, the Gaussian Mixture PHD filter \cite{Vo2006}, or GM-PHD filter, and the Sequential Monte Carlo PHD filter \cite{Vo2005}, or SMC-PHD filter.

As the objective is to incorporate the single-object filter, designed in the previous sections, into a multi-object framework, the choice of a Gaussian Mixture implementation of the PHD filter is the most appropriate. The transition $M^{i|j}_{t|t-1}$ is then the particle move between the disparity spaces $\hat{\bbD}_j$ and $\hat{\bbD}_i$, from time $t-1$ to time $t$. Note that the use of the Gaussian Mixture implementation requires additional assumptions:
\begin{enumerate}[label=\bfseries A.\arabic*,resume]
\item The probability of survival $p_S$ is state-independent.
\item \label{hyp:constantPd} The probability of detection $p_D$ is state-independent.
\end{enumerate}

With these assumptions it can be demonstrated~\cite{Vo2006} that the equations of the PHD filter propagate in closed form a Gaussian mixture of the form:
\eqns{
\mu^i_t(\bfy) = \sum_{k=1}^{N_t} w_k \calN(\bfy; \bfy^i_k, Q^i_k).
}
Note that the weight $w_k$ of the $k$\textsuperscript{th} term in the mixture does not depend on the space in which the Gaussian distribution is expressed.

However, Assumption \ref{hyp:constantPd} is too strong when considering a pair of cameras, as their field of view might, and will, significantly differ. It is then necessary to relax such an assumption and we discuss this in detail in the next section.

Following the choice of initialising the probability density 
with the first observation available, we adopt the observation-driven birth, detailed in \cite{Houssineau2010}. Note that previous attempts to use the PHD filter with cameras, e.g.\ \cite{Pham2007} or \cite{Laneuville2010}, required the scene to be bounded and/or the use of at least 3 cameras. These restrictions limit the impact of the error made when representing the uncertainty by a Gaussian distribution in $\hat{\bbD}$ and increase the observability of the objects as the problem of triangulation from 3 points of view is better constrained than from only 2.

\subsection{State-dependent probability of detection}
\label{ssec:state_dependent_detection}

As opposed to radar applications, the estimation of multiple objects from a camera pair requires the fields of view to be properly modelled. For this reason, Assumption \ref{hyp:constantPd} must be relaxed. Once again, we can resort to a solution similar to the particle move, introduced in the previous sections, in order to consider a state-dependent probability of detection.

\begin{figure}
\centering
\includegraphics[width=0.6\columnwidth]{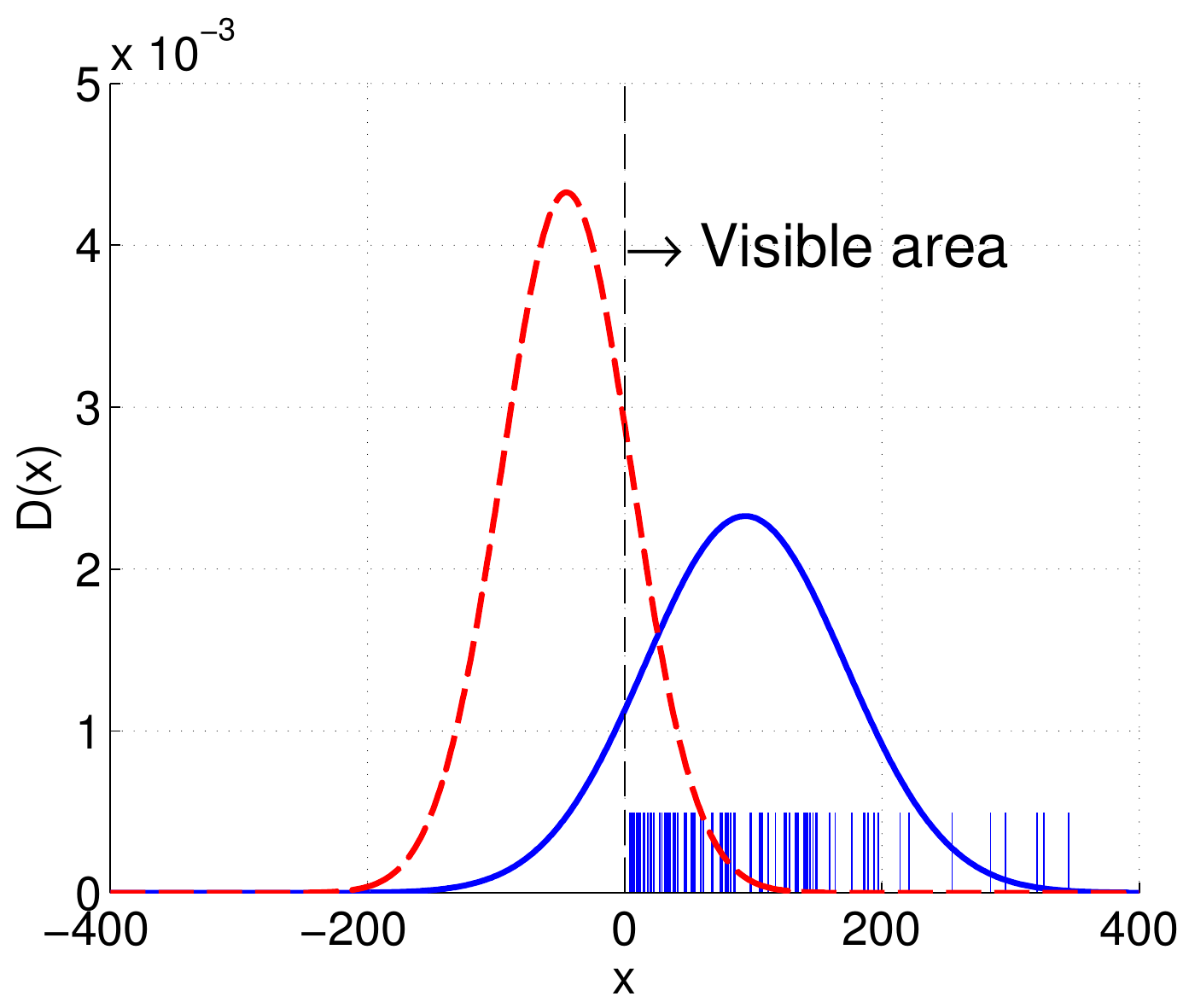}
\caption{Gaussian fitting for a state-dependent probability of detection in the $1$-D case. $p_D(\bfx) = 0.9$ for $\bfx \geq 0$ and $p_D(\bfx) = 0$ for $\bfx < 0$.
Solid line: detection term -- Dashed line: missed-detection term}
\label{fig:state_dependent_detection}
\end{figure}

Formally, the following two Gaussian distributions can be computed for each original Gaussian term in the mixture $\hat{\mu}^i_t$:
\begin{itemize}
\item one corresponding to the missed detection term:
\eqns{
w_{\circ,k} \calN(\bfy; \hat{\bfy}^i_{\circ,k}, \hat{Q}^i_{\circ,k}) \approx (1-p^i_D(\bfy)) w_k \calN(\bfy; \hat{\bfy}^i_k, \hat{Q}^i_k),
}
\item one corresponding to the detection term:
\eqns{
w_{\bullet,k} \calN(\bfy; \hat{\bfy}^i_{\bullet,k}, \hat{Q}^i_{\bullet,k}) \approx p^i_D(\bfy) w_k \calN(\bfy; \hat{\bfy}^i_k, \hat{Q}^i_k),
}
\end{itemize}
where the subscripts ``$\circ$'' and ``$\bullet$'' indicate missed detection and detection respectively. This is achieved by sampling particles according to the predicted law and then applying the state-dependent probability of detection before computing the mean and covariance of the obtained weighted set of particles. An example of such an approach is depicted in Figure \ref{fig:state_dependent_detection}. Denoting by $\hat{\mu}^i_{\circ,t}$ and $\hat{\mu}^i_{\bullet,t}$ the modified missed detection and detection first-moment densities, the PHD update can be expressed as
\eqnl{eq:PHDupdate}{
\mu^i_t(\bfy) = \hat{\mu}^i_{\circ,t}(\bfy)+\sum_{\bfz \in Z^i_t} \int \dfrac{L^i_t(\bfz|\bfy)\hat{\mu}^i_{\bullet,t}(\bfy)}
{\lambda^i c^i(\bfz) + \int L^i_t(\bfz|\bfy')\hat{\mu}^i_{\bullet,t}(\bfy') \d\bfy'}.
}

Note that the Gaussian distributions which depict objects that are almost surely inside or outside of the field of view can be kept as they are, so that only the weight $w_k$ changes. For instance, if the object is almost surely inside the field of view, $w_{\circ,k} = (1-p^{i,\inFOV}_D) w_k$ and $w_{\bullet,k} = p^{i,\inFOV}_D w_k$, where $p^{i,\inFOV}_D$ is the constant probability of detection within the field of view of the camera $C_i$.

Equipped with a suitable way of estimating multiple objects from a non-rectified camera pair, we now proceed to describe a solution for the problem of camera calibration in the next section.

\section{Camera calibration from multiple objects}
\label{sec:calibration}

Reliable estimation requires reliable knowledge of the sensor parameters, and thus sensor calibration has been a central problem in multi-object multi-sensor tracking. In the context of FISST, solutions to this problem have been derived recently~\cite{Lee2013,Mullane2011,Ristic2013}. However, these solutions have not been used for calibrating cameras. The objective of this section is to extend the multi-object estimation framework, described in the previous section, and present a method for calibrating a non-rectified camera pair by formulating a joint multi-object tracking and camera calibration algorithm.

\subsection{Model parameters}

The origin of the coordinate system is assumed to be aligned with the left camera position and orientation, so that only the right camera has to be calibrated in order to define the camera pair $(C_{\l},C_{\r})$. Let $\bbS_{\r} = \bbR^d$, $d > 0$, be the space in which the state of the right camera is described. In general, the components of a given state vector $\bfs$ in $\bbS_{\r}$ can be
\begin{itemize}
\item the camera's position in $\bbX$ ($3$-D),
\item the camera's orientation in $\bbX$ ($3$-D),
\item the velocity and rotation rates ($6$-D),
\item the focal length ($1$-D),
\item the coordinates of the principal point ($2$-D), and,
\item the image distortion ($1$-D) for a non-pinhole camera,
\end{itemize}
so that the dimension $d$ of the right-camera's state space can be as high as 16.

The objective is to jointly estimate the state of the multiple objects in the scene, as well as the state of the right camera, $C_{\r}$, relative to the left camera, $C_{\l}$. We thus introduce the joint probability distribution $\bfP^i_t$ which encompasses the right camera state $\bfs \in \bbS_{\r}$, as well as the multi-object state $Y$:
\eqns{
 \bfP^i_t(Y,\bfs) = P^i_t(Y|\bfs)p_t(\bfs),
}
where $p_t$ is a probability distribution over $\bbS_{\r}$. 

For the same reasons as the ones discussed in Section \ref{ssec:general_solution}, it continues to be impractical to work with multi-object densities directly, and the first-moment density
\eqnl{eq:calibphd}{
\bfmu^i_t(\bfy,\bfs) = \mu^i_t(\bfy|\bfs)p_t(\bfs)
}
is preferred. This relation holds as the first-moment density corresponding to a single-variate distribution is the distribution itself. Equation \eqref{eq:calibphd} indicates that the use of the PHD filter for propagating the first-moment density $\bfmu^i_t$ can be considered. We describe this approach in the next section.

\subsection{Conditional PHD filtering}

Due to the conditional nature of \eqref{eq:calibphd}, the derivation of the PHD filter results in an expression that is different to the usual PHD filter equations. The result of this derivation, detailed in \cite{Ristic2013}, can be expressed as
\eqns{
\bfmu^i_t(\bfy,\bfs) = \mu^i_t(\bfy|\bfs) \alpha_t(\bfs)\hat{p}_t(\bfs),
}
where $\mu^i_t(\cdot|\bfs)$ is found via the PHD update \eqref{eq:PHDupdate}, where $\lambda^i$, $c^i$, $L^i_t$ and $p^i_D$ might be dependent on $\bfs$, and where $\alpha_t(\bfs) \in [0,1]$ relates to the probability for the sensor state $\bfs$ to generate a successful multi-object update, expressed as
\eqns{
\alpha_t(\bfs) = \dfrac{\bfL^{\c}_t(Z^i_t|\bfs)}{\int \bfL^{\c}_t(Z^i_t|\bfs')\hat{p}_t(\bfs')\d\bfs'},
}
where $\bfL^{\c}_t(Z^i_t|\bfs)$, with ``c'' standing from ``calibration'', is interpreted as the likelihood of the observation set $Z^i_t$, given the camera state $\bfs$, defined as
\eqnsml{
\bfL^{\c}_t(Z^i_t|\bfs) = \exp\left(-\lambda(\bfs)-\int \mu^i_{\bullet,t}(\bfy|\bfs)\d\bfy \right)\\
\times\prod_{z \in Z^i_t} \left[\lambda(\bfs)c(z|\bfs) + \int L^i_t(z|\bfy,\bfs) \mu^i_{\bullet,t}(\bfy|\bfs)\d\bfy \right].
}
The expression of $\bfL^{\c}_t$ contains a product over the observations, assessing the probability for each of these to be either a spurious observation or to come from an object in $\calX_t$. This form confirms the status of a multi-object likelihood for $\bfL^{\c}_t$. 

Interestingly, the structure of the joint multi-object tracking and camera calibration is similar to the one derived for group tracking, see, e.g.,~\cite{Swain2010extended} and \cite{Swain2010first} . This similarity can be explained by the hierarchical structure shared by the two estimation problems.

As the single-object likelihood $L^i_t$ exhibits the same kind of non-linearity as the mapping from $\hat{\bbX}$ to $\hat{\bbD}_i$ or $\bbP_i$, we can readily conclude that the distribution $p_t$ is likely to be non-Gaussian in $\bbS_{\r}$. However, we do not wish to model the possibility for the right camera to be infinitely far from the left camera, and thus a particle representation is now suitable.

For these reasons, we select a particle representation of the camera distribution $p_t$, composed of $M_t$ particles $\{\bfs_k\}_{k=1}^{M_t}$, expressed as
\eqns{
 p_t(\bfs) \approx \sum_{k=1}^{M_t} \omega_k \delta_{\bfs_k}(\bfs),
}
where $\delta_{\bfs_k}$ is the Dirac function at point $\bfs_k$. The updated joint first-moment density $\bfmu^i_t$ can then be rewritten as
\eqns{
\bfmu^i_t(\bfy,\bfs) \approx \sum_{k=1}^{\hat{M}_t} \mu^i_t(\bfy|\hat{\bfs}_k) \alpha_t(\hat{\bfs}_k) \hat{\omega}_k \delta_{\hat{\bfs}_k}(\bfs),
}
so that  each possible camera predicted state $\hat{\bfs}_k$ is associated with a specific conditional first-moment density $\mu^i_t(\cdot|\hat{\bfs}_k)$, propagated with a GM-PHD filter.

In practice, particle implementations are known to be sensitive to the curse of dimensionality. The number of particles needed to maintain a certain approximation error grows exponentially with the number of state dimensions. Therefore, every effort should be made to decrease the number of calibration parameters being estimated. Rather than estimate all of the above mentioned parameters in one pass in a $16$-dimensional state space, we suggest the following approach:
\begin{enumerate}
 \item Assume that the camera pair is in a static configuration, in order to temporarily ignore the $6$ dimensions required for the motion estimation. The intrinsic parameters can then be estimated within a $10$-dimensional state space.
 \item Once the intrinsic parameters are known, the estimation of the position and velocity can then take place within a $12$-dimensional state space.
\end{enumerate}

\section{Results on simulated data}
\label{sec:simulated_data}

The proposed approach was validated with several simulated scenarios, depicting interesting examples of use. The basic experimental configuration is a pair of non-rectified cameras that observe a scene with objects that behave in different ways:
\begin{itemize}
  \item Single-object localisation
  \item Single-object tracking
  \item Multi-object tracking
  \item Camera calibration
\end{itemize}
These examples are presented in the following sections.

\subsection{Single-object localisation}
\label{ssec:single_object_localisation}

One of the strengths of the disparity space representation is that it allows for the definition of prior distributions, where a large range of distance values are taken into account, using a single Gaussian representation. This is advantageous for triangulation, since it limits the amount of resources that are necessary to define a prior distribution for a newly observed object, and then localise it using a Bayes update.

\subsubsection*{First scenario}

\begin{table}
\renewcommand{\arraystretch}{1.3}
    \caption{Simulated camera parameters}
    \label{tab:simulated_data:camera_params}
\centering
    \begin{tabular}{ r l | r l }
      Parameter & Value & Parameter & Value \\ \hline
      focal length $f$ & $-8\mm$ & principal point $u_0$ & $400$ \\
      pixel size $d_u$ & $8.9\mu\m$ & $v_0$ & $300$ \\
      $d_v$ & $9.0\mu\m$ & baseline $b$ & $1\cm$ \\
      observation noise $\sigma_u^2$ & $2$ & & \\
      $\sigma_v^2$ & $2$ & & \\ \hline \\
    \end{tabular}
\end{table}

\begin{figure}
  \centering
   \includegraphics[trim=180pt 315pt 180pt 320pt,clip,width=0.6\columnwidth]{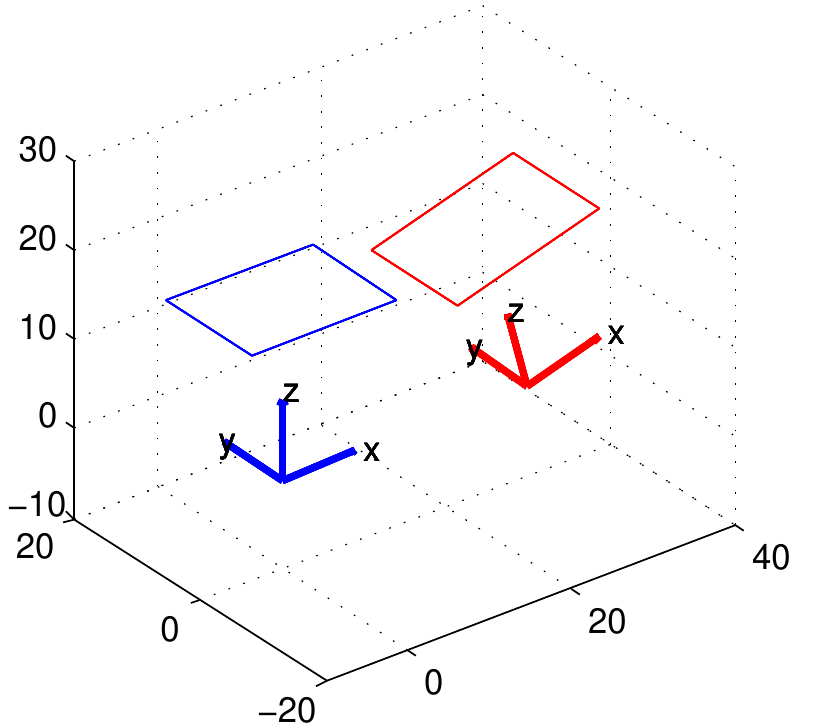}
  \caption{Configuration for the localisation experiment. The frames of reference of both cameras are shown.}
\label{fig:simulated_data:localisation_setup}
\end{figure}

In this scenario, the localisation capabilities of the proposed disparity space-based filter are compared against a particle filter. Two cameras are set up as in Figure \ref{fig:simulated_data:localisation_setup}, i.e., the first camera is at the centre of the coordinate system and the second camera is translated by $30\,\cm$ along the $x$ axis with respect to the first, and rotated $\pi/12$ radians around the $y$ axis. The objects are located along the $z$ axis at different distances, and are observed by the modelled pinhole cameras. The camera parameters can be found in Table \ref{tab:simulated_data:camera_params}. The particle filter was run with 100 particles which were sufficient to cover a ray through the region of interest. The proposed solution also uses 100 particles for the particle move step.

\begin{table}
\renewcommand{\arraystretch}{1.3}
    \caption{Mean and covariance of the RMSE for single object localisation, for different parameters for the prior distribution. (DS: Disparity space, PF: Particle Filter)}
    \label{tab:simulated_data:localisation}
\centering
    \begin{tabular}{r c c c}
      Case &    $\mu_d=6, \sigma_d^2=4$ & $\mu_d=7, \sigma_d^2=5.4$ & $\mu_d=8, \sigma_d^2=7.1$ \\ \hline
      $0.5\m$ (DS) & 1.11 (4.30)   & 0.462 (0.291) & 0.552 (0.544) \\
      $0.5\m$ (PF) & 156 (250)     & 94.5 (62.8)   & 62.9 (52.03)  \\
      $1\m$ (DS) & 0.845 (0.630) & 0.998 (0.667) & 1.23 (1.09)   \\
      $1\m$ (PF) & 1.30 (0.98)   & 1.11 (0.686)  & 1.03 (0.669)  \\
      $1.5\m$ (DS) & 1.80 (1.63)   & 1.88 (1.48)   & 1.78 (1.35)   \\
      $1.5\m$ (PF) & 2.35 (1.69)   & 2.81 (1.95)   & 2.23 (1.92)   \\ \hline \\

    \end{tabular}
\end{table}

The state estimator used for the particle filter is the MAP of the distribution in $\bbX$, while the state estimator of the disparity space filter is taken as the MAP of the distribution in disparity space, mapped to $\bbX$. 50 Monte Carlo runs were performed, where the objects were localised by both filters, using different values for the prior distribution, and results can be seen in Table \ref{tab:simulated_data:localisation}, where the MAP estimate of both filters is compared to the ground truth to compute the Root Mean Square Error (RMSE). The different values of the prior were selected so that the region of interest would be sufficiently covered by particles. The proposed filter performs better than the particle filter in most cases, but it is notable how dependent the particle filter is to a proper initialisation, which requires {\em a priori} knowledge about the object's depth.

\subsubsection*{Second scenario}
In this scenario, the performance in localisation of the disparity space-based solution is compared against inverse depth, as in~\cite{Houssineau2012}. The configuration of the camera pair is similar to the one used in the previous scenario, except that the right camera is $80\,\cm$ away from the left camera and is rotated by an angle of $-\pi/4$ radians. The target is also located on the $z$ axis, $150\,\cm$ away from the left camera. The initialisation of the inverse depth component is made equivalent to the one used for disparity, with $\mu_d = 12.5$ and $\sigma_d = 4.2$.

\begin{figure}
  \centering
   \includegraphics[trim=170pt 350pt 175pt 350pt,clip,width=0.8\columnwidth]{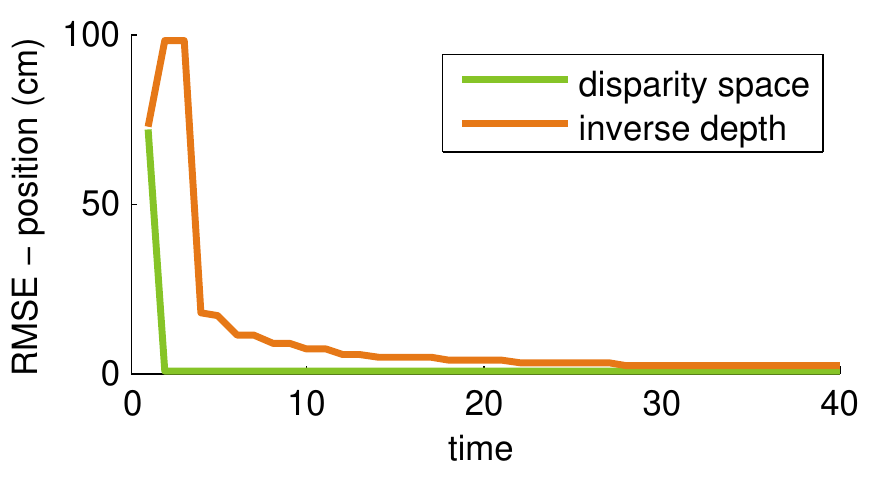}
  \caption{Performance of the estimation in disparity space compared with inverse depth for the localisation of a static object with a non-rectified pair of cameras.}
\label{fig:simulated_data:localisation_results}
\end{figure}

The average performance over 100 Monte Carlo runs for the two localisation algorithms is shown in Figure \ref{fig:simulated_data:localisation_results}. It appears that the inverse depth approach does not cope well with the non-linearity of the observation function and makes a significant error at the second time step, whereas the disparity space approach manages to localise the target almost instantly. This result can be explained by the limitations of the EKF when dealing with non-linear functions such as the one depicted in Figure \ref{fig:nonLinearity}. This example corroborates the fact that the proposed solution does manage to propagate the uncertainty between the left and right disparity spaces, $\bbD_{\l}$ and $\bbD_{\r}$, as already suggested in Figure \ref{fig:UKF_particle}.

\subsection{Single-object tracking}
\label{ssec:single_object_tracking}

\begin{figure}
\centering
\includegraphics[trim=145pt 260pt 150pt 260pt,clip,width=0.8\columnwidth]{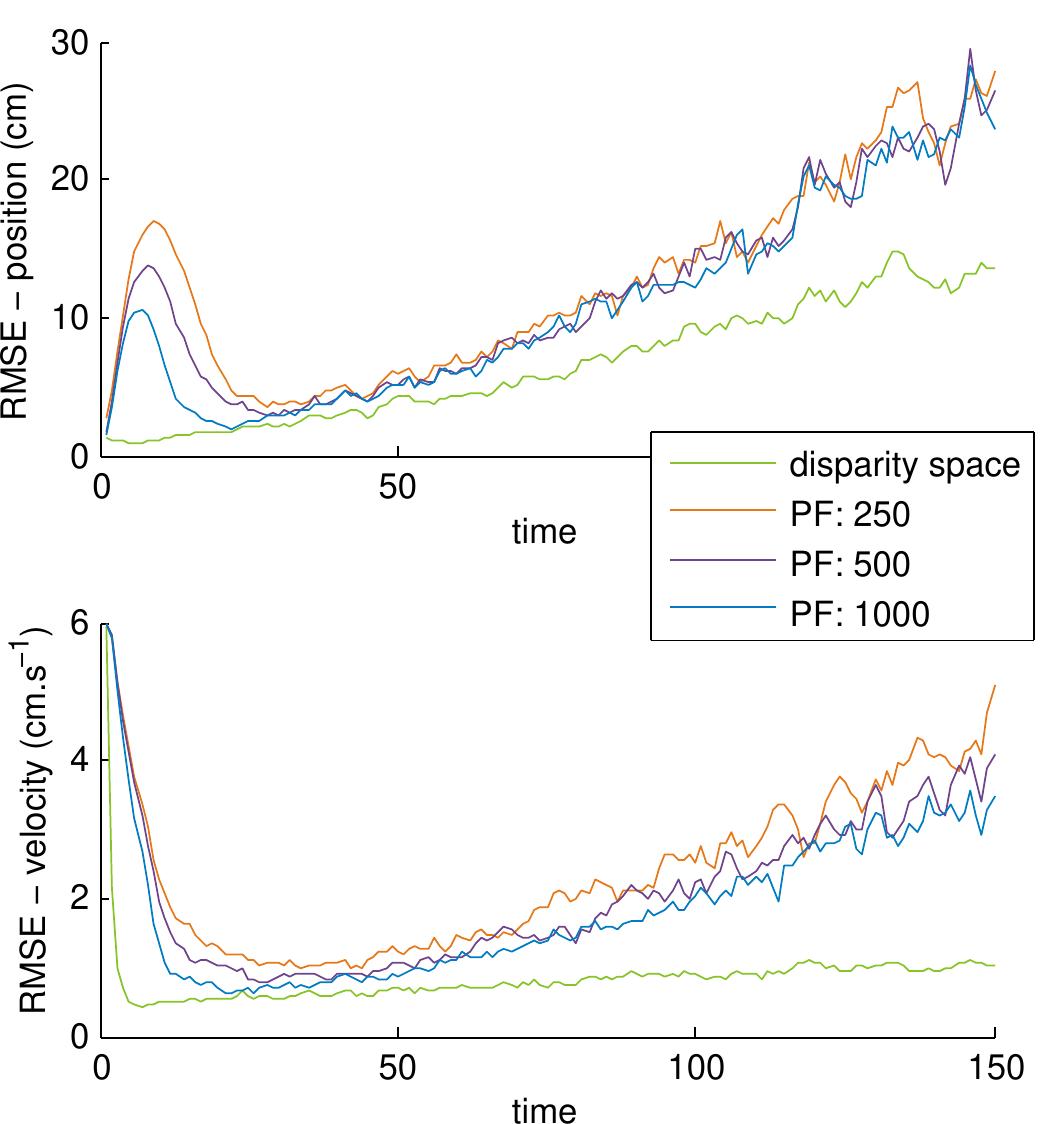}
\caption{Performance of the proposed single-object tracking algorithm (disparity space) compared against a particle filter with 250, 500 and 1000 particles (PF:250, PF:500 and PF:1000).}
\label{fig:simulated_data:single_tracking_results}
\end{figure}

The suitability of the disparity space parametrisation for tracking was evaluated through an experiment where an object moves away from the left camera with a nearly-constant velocity. This was done to analyse how capable the parametrisation is to deal with smooth changes in distance. The configuration of the camera pair is as follows: the left camera $C_{\l}$ is at $(-20,0,0)$ and is rotated by an angle of $\pi/12$ about the $y$ axis, while the right camera $C_{\r}$ is at $(20,0,0)$ and is rotated by an angle of $-\pi/12$ about the $y$ axis. As before, the filter was initialised with a prior distribution in disparity space, and then it was successively updated with measurements that were acquired synchronously from both cameras. The experiments consisted of tracking an object with an initial velocity of $6\,\cm.\s^{-1}$ along the $z$ axis.

The performance of the proposed solution is once again compared against a particle filter. The objective is to demonstrate that the approximation made when fitting a Gaussian distribution after the particle move is compensated by the gain in accuracy obtained during the observation update. The average performance over 100 Monte Carlo runs for the disparity space approach is shown in Figure \ref{fig:simulated_data:single_tracking_results}, together with the performance of a particle filter for different numbers of particles. It appears that the particle filter struggles at the initialisation, both in the estimation of position and of the velocity of the target. This can be explained by the degeneracy of the set of particles when the prior distribution is updated by the observation from the right camera. Towards the end of the experiment, the target is up to $10\,\m$ away from the camera pair, and the estimation is once again made difficult for the particle filter, as resampling is needed more and more frequently to cope with the noise in the observations. The disparity space approach only uses $250$ particles for the particle move and does not require resampling to be applied. As a consequence, the computational time is equivalent to a particle filter with $250$ particles, while the performance is better than that of a particle filter with $1000$ particles.

\subsection{Multi-object tracking}
\label{ssec:multi_object_tracking}

Having assessed the performance of the disparity space representation for tracking a single target, an experiment was done to evaluate the performance of a PHD filter equipped with the disparity space representation for simultaneously tracking multiple objects. To evaluate the performance of the multiple target tracker, the OSPA metric \cite{Schuhmacher2008} was utilised. This metric is commonly employed to measure the performance of multi-object tracking filters. It gives the distance between two sets of points by first solving the optimal assignment problem and returning a weighted combination of the average distance between the matched points and the difference in cardinality between the two sets. In more specific terms, the distance is formulated as follows. Let $d^{(c)}(x,y):=\min \left( c,\Vert x-y\Vert \right)$ for $x,y\in \bbX$, and $\Pi _{k}$ denote the set of permutations on $\{1,2,\ldots ,k\}$ for any positive integer $k$. Then, for $m \leq n$, $p\geq 1$, and $c>0$, the OSPA distance between two finite sets $X=\{x_{1},\ldots ,x_{m}\}$ and $Y=\{y_{1},\ldots ,y_{n}\}$ of points in $\bbX$ is
\eqns{
\bar{d}_{p}^{(c)}(X,Y):=
\Bigg[ \frac{1}{n}\bigg[ \min_{\pi \in \Pi _{n}}\sum_{i=1}^{m}d^{(c)}(x_{i},y_{\pi(i)})^{p}+c^{p}(n-m)\bigg]\Bigg]^{\frac{1}{p}}
}
For the case of $m = n = 0$, $\bar{d}_{p}^{(c)}(X,Y) = 0$, and for $m > n$, $\bar{d}_{p}^{(c)}(X,Y) = \bar{d}_{p}^{(c)}(Y,X)$. Finding the minimum permutation $\pi \in \Pi_k$ is equivalent to solving the optimal assignment problem with cost function $d^{(c)}(x,y)$. The second term in the summation is the penalisation for mismatched cardinalities. The cutoff parameter $c$ serves to determine the relative contributions of localisation and cardinality errors, where a larger value shifts the emphasis towards the latter. The order parameter $p$ influences the metric's sensitivity to outlier points. The distance function $d$ was selected as the Mahalanobis distance.

\begin{figure}
\centering
\includegraphics[trim=165pt 375pt 180pt 375pt,clip,width=0.8\columnwidth]{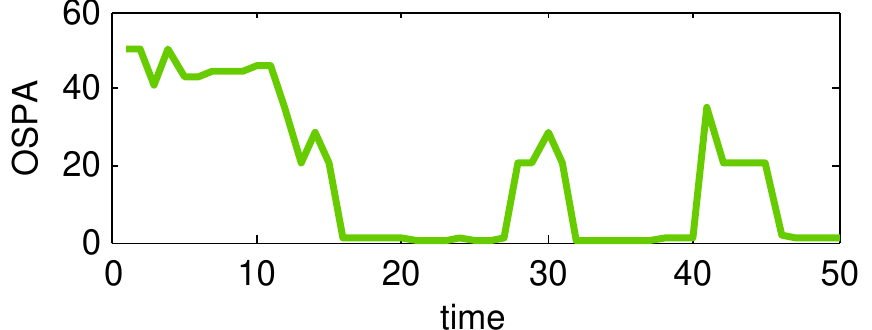}
\caption{Performance of the multi-object tracking algorithm.}
\label{fig:simulated_data:tracking_results}
\end{figure}

For this experiment, the cameras were set up in a configuration similar to the one considered in Section \ref{ssec:single_object_tracking}, i.e., the cameras were located on the $y=z=0$ plane at $-20\,\cm$ and $20\,\cm$ along the $x$ axis, and rotated $\pi/12$ and $-\pi/12$ radians around the $y$ axis, respectively.
The model parameters used for the GM-PHD filter are detailed in Table \ref{tab:model_parameters}. Six objects moving according to a constant velocity model were observed by the two cameras. In Figure \ref{fig:simulated_data:tracking_results}, the evolution of the OSPA metric can be seen, showing good agreement between the ground truth and the obtained estimates. In this figure, it appears that the OSPA distance increases around times $30$ and $40$. This phenomenon can be explained by crossing bearings which induce the momentary loss of one track.

\subsection{Camera calibration}
\label{ssec:camera_calibration_results}

\begin{figure}
\centering
\includegraphics[trim=165pt 310pt 180pt 310pt,clip,width=0.8\columnwidth]{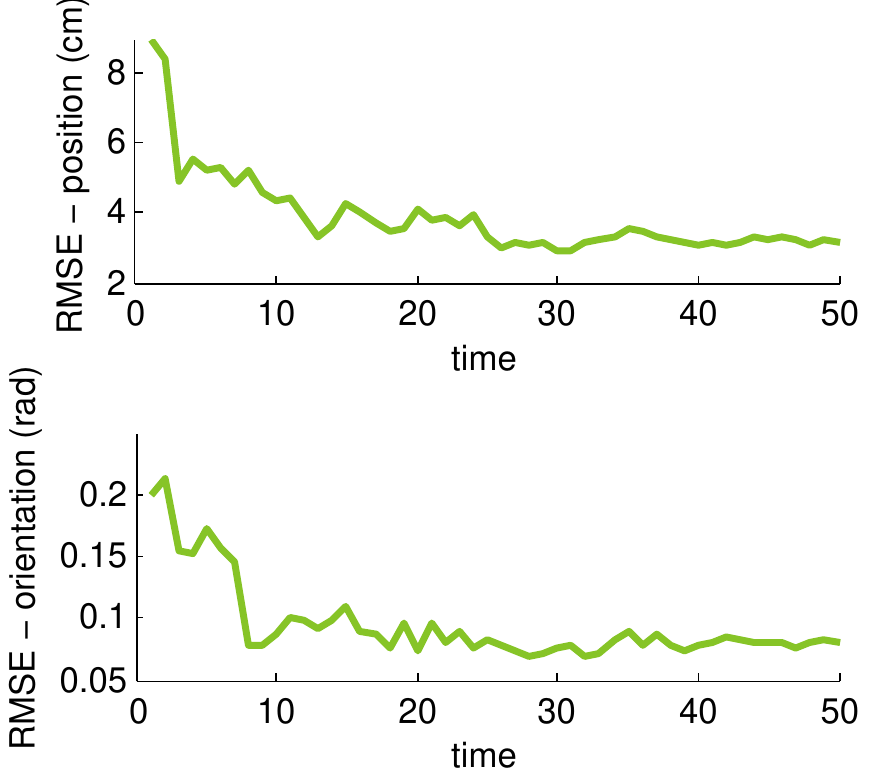}
\caption{Average performance over 20 Monte Carlo runs for the estimation of the extrinsic parameters of the right camera for the proposed joint multi-object tracking and camera calibration algorithm.}
\label{fig:simulated_data:calibration_results}
\end{figure}

\begin{table}
\renewcommand{\arraystretch}{1.3}
    \caption{Model parameters for the GM-PHD filter}
    \label{tab:model_parameters}
\centering
    \begin{tabular}{ r|l}
      Parameter & Value \\ \hline
      merging distance & $7$ \\
      pruning threshold & $10^{-6}$ \\
      false alarm Poisson parameter $\lambda^i$ & $10$ \\
      probability of detection $p^i_D$ & $0.95$
    \end{tabular}
\end{table}

In sections \ref{ssec:single_object_localisation} and \ref{ssec:single_object_tracking}, the proposed solution was shown to outperform inverse depth-based algorithms for the localisation of a static target, and a particle filter for the tracking of a moving target, both from non-rectified cameras. Then, in Section \ref{ssec:multi_object_tracking}, the multi-object tracking performance of the proposed solution was evaluated. In this section, the objective is to demonstrate that the extrinsic parameters of the right camera can be estimated by tracking $6$ non-cooperative moving targets.

For this experiment, the cameras were set up in a configuration similar to the one considered in Section \ref{ssec:multi_object_tracking}. Figure \ref{fig:simulated_data:calibration_results} shows the convergence of the estimation in position and orientation for a 6-D calibration problem with the following parameters:
\begin{itemize}
\item Number of particles for calibration: 1500,
\item Prior position uncertainty:
\eqns{
\sigma_x = \sigma_y = \sigma_z = 5\,\cm,
}
\item Prior orientation uncertainty:
\eqns{
\sigma_{\phi} = \pi/24 \quad\mbox{and}\quad \sigma_{\theta} = \sigma_{\psi} = \pi/180.
}
\end{itemize}

Figure \ref{fig:simulated_data:calibration_results} demonstrates that the proposed solution enables the calibration of non-rectified cameras from multiple, non-cooperative, moving objects, when the data association is not known. To the best of the authors' knowledge, there is no existing alternative solutions for such a problem. Note that the values of the standard deviation in position are large enough to set up the initial value by the naked eye, as it covers a $30\,\cm$ error in each direction, whereas the actual distance between the two cameras is $40\,\cm$. The uncertainty for the orientation $\phi$ around the $y$-axis is also relatively large, as it covers up to $45^{\circ}$ uncertainty. The orientations $\theta$ and $\psi$ around the $x$ and $z$ axis, respectively, are assumed to be better known, with only $6^{\circ}$ of coverage for these components. A higher number of particles would be required to allow for a higher uncertainty.

\section*{Conclusion}

This paper addresses the problem of statistical estimation from camera images, and proposes novel methods for triangulation, feature correspondence, and camera calibration, based on a parametrisation related to pixel coordinates and a method of sensor registration from measurements of moving targets on different sensors. The method exploits the use of disparity space for statistical estimation from rectified and non-rectified camera networks and develops a new method of triangulation from multiple cameras that can track objects. The method is compared with the nonlinear variants of the Kalman filter and particle filter in different scenarios and is shown to give better performance. Moreover, the choice of parametrisation is shown to provide more stable results than the inverse depth parametrisation. 

The approach is extended for estimating multiple objects using the Probability Hypothesis Density (PHD) filter, both for static objects, which provides a means of automatic feature correspondence estimation, and for dynamic objects. The approach inherits the advantages of the Finite Set Statistics framework, i.e., automatic discrimination between targets and clutter, estimating the correct number of objects, and automatic identification of new objects. The approach is then further extended to develop a method for calibrating cameras that does not require advance knowledge of feature correspondences or that the cameras are synchronous. The approach is based on a recent Bayesian approach for sensor registration, called the single-cluster PHD filter. The performance of the calibration method is shown in statistical simulations and compared with a scenario with a known calibration.

\section*{Acknowledgement}

Jeremie Houssineau has a PhD scholarship sponsored by DCNS and a tuition fees scholarship by Heriot-Watt University. This work was supported by a Royal Academy of Engineering/EPSRC Research Fellowship and the Engineering and Physical Sciences Research Council grant EP/J012432/1.

\bibliographystyle{abbrv}
\bibliography{calibration,proposal-camerafusion2}

\end{document}